\documentclass[sn-mathphys,Numbered]{sn-jnl}% Math and Physical Sciences Reference Style
\usepackage{graphicx}%
\usepackage{multirow}%
\usepackage{amsmath,amssymb,amsfonts}%
\usepackage{amsthm}%
\usepackage{mathrsfs}%
\usepackage[title]{appendix}%
\usepackage{xcolor}%
\usepackage{textcomp}%
\usepackage{manyfoot}%
\usepackage{booktabs}%
\usepackage{algorithm}%
\usepackage{algorithmicx}%
\usepackage{algpseudocode}%
\usepackage{listings}%
\usepackage{natbib}

\usepackage[english]{babel}
\usepackage{fixltx2e}
\usepackage{setspace}
\usepackage{array}
\usepackage{tabularx}
\usepackage{calc}
\usepackage{makecell}

\usepackage{subcaption}

\theoremstyle{thmstyleone}%
\theoremstyle{thmstyletwo}%

\theoremstyle{thmstylethree}%

\begin{document}

\title[Article Title]{T-Code: Simple Temporal Latent Code for Efficient Dynamic View Synthesis}

%%=============================================================%%
%% Prefix	-> \pfx{Dr}
%% GivenName	-> \fnm{Joergen W.}
%% Particle	-> \spfx{van der} -> surname prefix
%% FamilyName	-> \sur{Ploeg}
%% Suffix	-> \sfx{IV}
%% NatureName	-> \tanm{Poet Laureate} -> Title after name
%% Degrees	-> \dgr{MSc, PhD}
%% \author*[1,2]{\pfx{Dr} \fnm{Joergen W.} \spfx{van der} \sur{Ploeg} \sfx{IV} \tanm{Poet Laureate} 
%%                 \dgr{MSc, PhD}}\email{iauthor@gmail.com}
%%=============================================================%%

\author[1]{\fnm{Zhenhuan} \sur{Liu}}\email{pingfan\_zhilu\_law268@sjtu.edu.cn}

\author[1]{\fnm{Shuai} \sur{Liu}}\email{shuailiu@sjtu.edu.cn}

\author*[1]{\fnm{Jie} \sur{Yang}}\email{jieyang@sjtu.edu.cn}
\author*[1]{\fnm{Wei} \sur{Liu}}\email{weiliucv@sjtu.edu.cn}

\affil*[1]{\orgdiv{Department of Automation}, \orgname{Shanghai Jiao Tong University}, \orgaddress{\street{Dongchuan Road}, \city{Shanghai}, \postcode{100190}, \state{State}, \country{Country}}}

%%==================================%%
%% sample for unstructured abstract %%
%%==================================%%

\abstract{
Novel view synthesis for dynamic scenes is one of the spotlights in computer vision. The key to  efficient dynamic view synthesis is to find a compact representation to store the information across time. Though existing methods achieve fast dynamic view synthesis by tensor decomposition or hash grid feature concatenation, their mixed representations ignore the structural difference between  time domain and  spatial domain, resulting in sub-optimal computation and storage cost. This paper presents T-Code, the efficient decoupled  latent code for the time dimension only. The decomposed feature design  enables customizing modules to cater for different   scenarios with individual specialty and yielding desired results at lower  cost.  Based on T-Code,  we propose our highly compact hybrid neural graphics primitives (HybridNGP) for multi-camera setting and deformation neural graphics primitives with T-Code (DNGP-T) for monocular scenario. Experiments show that HybridNGP delivers high fidelity results at top processing speed  with much less storage consumption, while DNGP-T achieves state-of-the-art quality and high training speed for monocular reconstruction. 
}


%%\pacs[JEL Classification]{D8, H51}

%%\pacs[MSC Classification]{35A01, 65L10, 65L12, 65L20, 65L70}

\maketitle

\section{Introduction}\label{sec1}

%  Novel View synthesis problem
Visual perception is a crucial component of artificial intelligence systems. 
 Novel view synthesis has been one of the  spotlights in computer vision. Novel view synthesis is defined as predicting a new observation of a 3D scene upon the given observed images, usually by regressing the relationship from the known camera poses to existing observations. Neural Radiance Field  \cite{nerfeccv} (NeRF) leverages differentiable volume rendering and  outperforms previous methods  \cite{neuralvolume,llff} by a remarkable margin.

Numerous methods have been proposed to improve and extend NeRF, including to enhance the reconstruction quality by frustum sampling   \cite{Mipnerf}, to improve the robustness against distractors   \cite{robustnerf}, to alleviate the urge need for data   \cite{regnerf,dsnerf} and to accelerate the reconstruction  and rendering process   \cite{plenoxels,kilonerf,recursivenerf}. Thanks to the previous efforts and the benefits from structure-from-motion packages like COLMAP  \cite{sfmotion}, researchers are able to reconstruct a 3D scene within minutes  \cite{ingp} upon simple image observations, or render new observations  at real-time speed on ordinary consumer-level laptops  \cite{yariv2023bakedsdf}, or even edit the geometry of the scenes  \cite{editnerf}. Besides the work on synthesizing tiny-scale scenes, recent work on large scale outdoor datasets  \cite{mipnerf360,F2nerf,suds} also shows the viability of the volume rendering and the potential to achieve virtual reality by NeRF.

% 没地方放了，这一段感觉可以不要了
%To be specific, given an observation, \emph{i.e.} an image of the 3D scene and the pose of the camera, NeRF samples a set of points along the camera ray, and uses heavy MLPs to regress the relationship from the Fourier-encoded coordinates and the observing direction to the optical attributes, \emph{i.e.} the color $\mathbf{c} $ and the volume density $\sigma$ of the points. With $\sigma$ and $\mathbf{c}$, a 2D image can be predicted through the sampled 3D representation via volume rendering.

%--------------------------------------------- Figure ------------------------------------
\begin{figure}[t]
    \centering
    \includegraphics[width=0.8\textwidth]{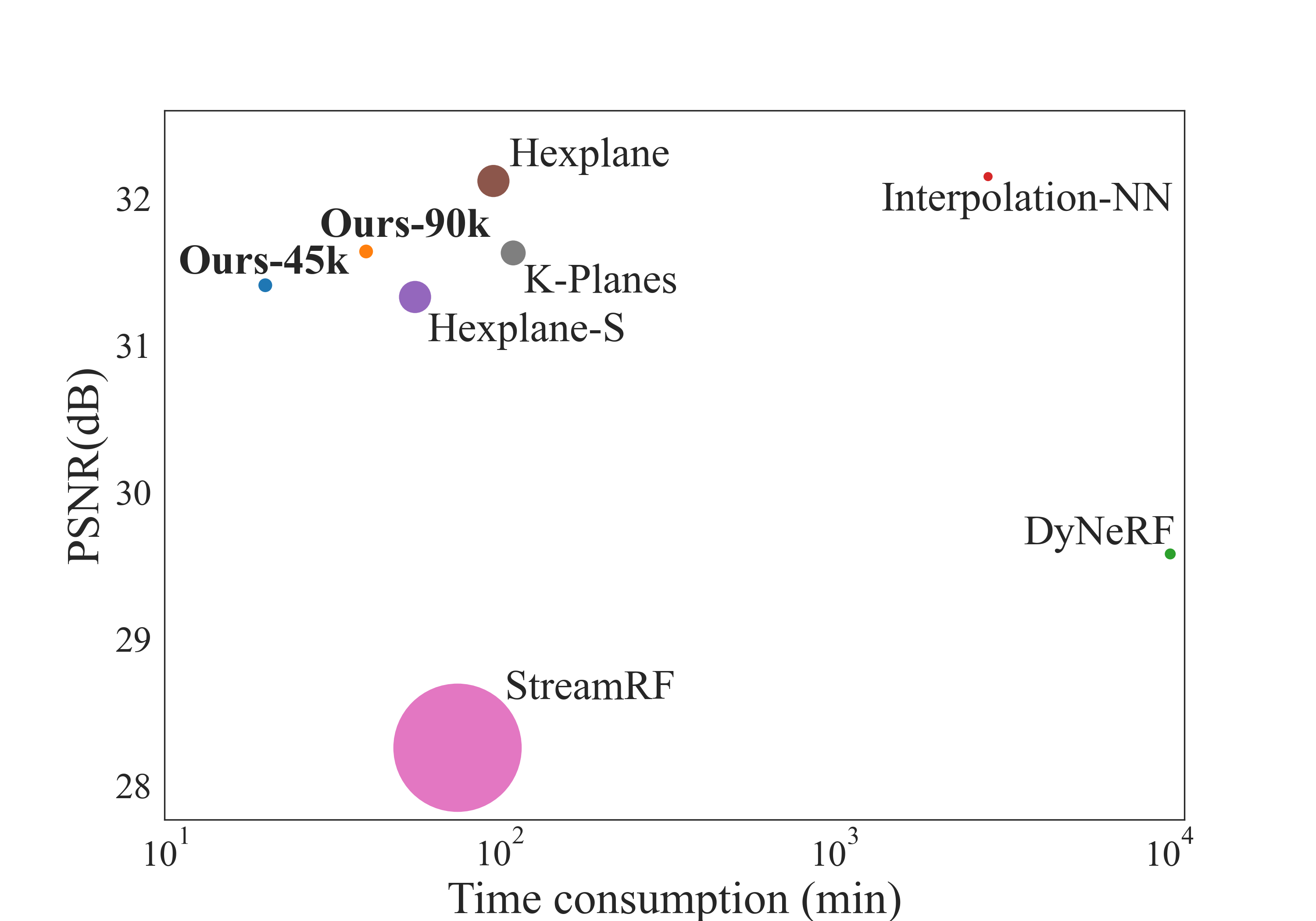}
    \caption{Visualization of quantitative comparisons between the proposed HybridNGP and other high-fidelity reconstruction methods on DyNeRF dataset. Sizes of the markers denote the storage consumption of different models. Benifiting from the efficiency of T-Code, our HybridNGP achieves fast and high-fidelity reconstruction at very low storage  cost.}
    \label{fig:quantile}
\end{figure}
%--------------------------------------------- -----------------------------------

%--------------------------------------------- Figure ------------------------------------
\begin{figure}[htbp]
    \centering
    \includegraphics[width=0.8\textwidth]{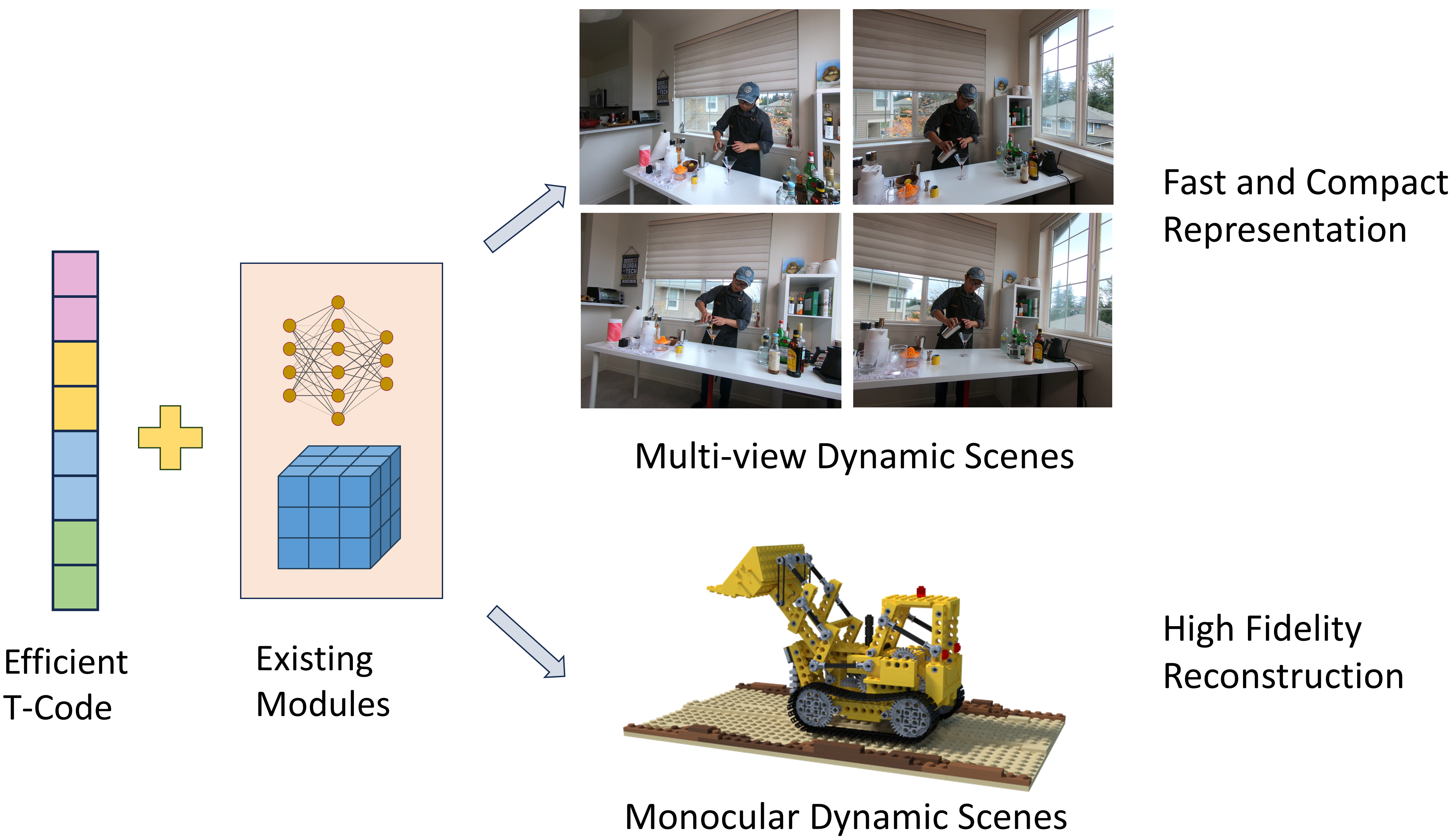} % Reduce the figure size so that it is slightly narrower than the column.
    \caption{Illustration of our temporal encoding T-Code, which serves as a compact representation for temporal features.  It can be combined with differently structured representations for dynamic view synthesis in varied scenarios with advantages in speed, compactness or quality. }
    \label{fig:outline}
\end{figure}
 %-------------------------------------------------------------------------------   

%hash grids 
Fast  3D scene reconstruction poses the need for compact and disentangled representation. This insight comes from the fact that the most space in an arbitrary 3D scene is vacuum, and the features at different locations may be decoupled. Based on this fact, previous researchers replace the large MLP in NeRF with varied representations for efficient reconstruction. Currently, the most effective representations for reconstructing a 3D scene are hash grids   \cite{ingp} and tensor decomposition   \cite{tensorf}. Both of these methods achieve  decoupled and parallel representation for 3D reconstruction at  high processing speed.

% novel view synthesis problem in dynamic scenes.
In contrast to the glorious accomplishments for static scenes, there remain difficulties for efficient reconstruction of dynamic scenes. In general, dynamic view synthesis, especially  in a fast and storage-friendly manner,  is difficult in three aspects. The first is  to  compress the temporal variance  across frames into a compact framework at minimum confusion across the time axis. As noted by previous researches, the mechanical motion, appearance variance and the topological changes are difficult to model and naive representations  may bring significant performance downgrade  \cite{nsff, Dnerf,Dynerf}. The second  is to cope with the structural difference. High quality dynamic view synthesis requires a precise representation in both  spatial and time domain. However,  learning the ambiguous structure of the sole time dimension is difficult,  let alone learning a unified representation in four diversely-structured dimensions. To cope with the four axes, current methods tend to increase the complexity of the representation   \cite{HexPlane_,kplanes, tineuvox, park2023temporal} or learn them unitedly with large neural modules   \cite{Dnerf,Dynerf,park2023temporal}.
However, complex models remain a burden in both computation and storage, leaving space for efficient reconstruction  with less parameters. The third is the variance of data. Currently, there are multi-camera datasets like DyNeRF  \cite{Dynerf}, and mono-camera datasets like D-NeRF   \cite{Dnerf} readily available. Since data are acquired under differed settings, it is difficult to form a unified problem structure.

Based on the analysis above, we introduce our compact temporal latent codes, \emph{i.e.} the T-Code for decomposed modeling of transient features.  As illustrated in figure~\ref{fig:outline}, we prove the viability of T-Code under different scenarios.  Combining T-Code with spatial hash encoding, we formulate our highly compact hybrid neural graphics primitives (HybridNGP), which achieves top reconstruction quality at  high speed for multi-camera dynamic scene synthesis, as compared with previous methods in figure~\ref{fig:quantile}. Our T-Code can also be combined with deformation-based hash encoding and yield state-of-the-art results under mono-camera setting.
The main contributions of the paper are as follows:
\begin{itemize}
    \item We are the first to use a compact 1D encoding (T-Code) in time domain for
    general dynamic scene reconstruction, explicitly disentangling the temporal features with the spatial ones. 
    
    \item We combine the T-Code with simple spatial hash encoding to cater for   multi-camera scenarios. Our method is able to achieve high fidelity reconstruction at much faster speed than recent approaches, while the storage consumption is also much smaller.
        
    \item We also combine T-Code with  deformation-based hash encoding  to cater for mono-camera scenarios. Experiments show that  the united representation yields state-of-the-art reconstruction quality with the merits of T-Code.
\end{itemize}

\section{Related Work}
\subsection{Acceleration Methods for Novel View Synthesis.}

As aforementioned, accelerated NeRF reconstruction usually requires a compact and  decoupled representation. The decomposition may be achieved by representing scenes as independent voxel grids  \cite{plenoxels}. However, 3D scenes consists of sparse features with different frequency patterns:  high-resolution representation at any location may result in storage and computation stress, while low-resolution representation can lead to quality downgrade.

 Instant-NGP  \cite{ingp} utilizes multi-resolution hash encoding to solve the dilemma, and they adopt a tiny  MLP to learn the fusion  of multi-level features. 
Their decoupled spatial  hash grids serve as a parallel representation, significantly accelerating the training step. 
Another paradigm for fast novel view synthesis is by tensor decomposition like TensoRF  \cite{tensorf}.  Representing the voxel grids as vector-matrix products, TensoRF achieves fast rendering and high quality with high tensor resolution.

\subsection{Dynamic Novel View Synthesis}
D-NeRF  \cite{Dnerf} blazes the trail for dynamic scene reconstruction.  They construct a synthesized dataset, in which the simulated dynamic scenes are observed through a single camera.    D-NeRF uses a large deformation MLP to encode the motion trajectory of points,  mapping every queried $(x,y,z)$ point in time $t$ to $(x',y',z')$ to feed into downstream NeRF MLP. With  Fourier-encoded time and spatial coordinates and the global perception of large MLP, D-NeRF gains a significant advantage over naive NeRF with time  \cite{Dnerf} but at a high computational cost.

TineuVox  \cite{tineuvox} replaces the geometry MLP in D-NeRF  with dense voxel grids to decouple the local spatial information. They represent the motion and transient effects by a tiny deformation MLP, and they further adopt multi-distance interpolation to fuse local and global patterns. TineuVox achieves high reconstruction quality and fast training on D-NeRF dataset. 
They adopt fourier temporal encoding  and MLP decoder for temporal information, and find   it less useful than the deformation MLP in the ablation study. 

DyNeRF  \cite{Dynerf} is the first to achieve high fidelity dynamic view synthesis in real scenes.  Conducting experiments on their multi-view video dataset, they point out that simply adding a encoded dimension into naive NeRF is not satisfactory. Instead, they encode the temporal information into a 1024-dim time-variant latent code. Hierarchical importance sampling is performed to train a united neural representation for the dynamic  scene. Although DyNeRF proves the viability of independent representation in the time axis,  reconstruction  of  their heavy neural representation of a 10-second dynamic scene  consumes thousands of GPU hours. To reduce GPU consumption and achieve incremental learning,  SteamRF  \cite{streamrf} adopts explicit voxel representation. They achieve on-the-fly training of DyNeRF, but at quite large storage cost.

Grid-interpolation   \cite{park2023temporal} is another grid-based method for fast dynamic view synthesis. Deeming the interpolation is crucial for dynamic view synthesis, they utilize the interpolation in both 3D and 4D volume. They concatenate the 3D hash grid features with  4D hash grid features and yield fast dynamic view synthesis on D-NeRF dataset. However, due to  the mixed hashing process  and confusion in time domain and spatial domain \cite{ingp,park2023temporal}, their 4D hash features do not produce top reconstruction quality. ReRF  \cite{rerf} also shows that simple 4D hash features are not enough for high-quality dynamic view synthesis. Sharing the same insight with Grid-interpolation  \cite{park2023temporal},  Interpolation-NN  \cite{park2023temporal} is a solution for high fidelity multi-camera video synthesis, yielding comparable results with DyNeRF at a smaller storage and computation cost.

HexPlane  \cite{HexPlane_} introduces fast tensor representation to dynamic view synthesis. By reducing the 4D volume to  products of 6 sub-planes, HexPlane achieves compactness and fast training. However, their symmetric representation ignores the internal structure differences in different axes, consuming almost 10 times of the storage of DyNeRF. The recent  K-Planes  \cite{kplanes} also shares similar insights and drawbacks with HexPlane.

\section{Method}

\subsection{Problem Formulation}

Regarding novel view synthesis as a machine learning problem, the key is to find a proper structure for the relationship between the  camera pose and  image observation. Following recent methods    \cite{Dynerf,kplanes,HexPlane_}, we solve the problem in a view of 3D geometry estimation. Considering each pixel of the input 2D image as the reduced observation of the 3D points, we train the representation along the camera ray to reconstruct the 3D scene where the camera rays are defined as:
%---------------------------Equation-----------------------
\begin{equation}
    \mathbf{r}(u)=\mathbf{o}+u \mathbf{d},
     \label{cameraray}
\end{equation}
%-----------------------------------------------------------
where $\mathbf{o}$ is the optical center of the camera, $\mathbf{d}$ is the direction off the camera. For each  point along the camera ray with coordinates $(x,y,z)$ observed at direction $\mathbf{d}=(\phi,\psi)$,  a field function $F$ is queried  to obtain the optical attributes $(\sigma, \mathbf{c})$. Conditioned on time $t$, the entire model is defined as:
%---------------------------Equation-----------------------
\begin{equation}
     (\sigma, \mathbf{c})=F_{\Theta}(x,y,z, t ,\phi,\psi),
     \label{nerfquery}
\end{equation}
%-----------------------------------------------------------
where the function $F$ denotes the dynamic NeRF model with varied structures, and $\Theta$ denotes trainable model parameters. Following previous work  \cite{nerfeccv}, our method apply a two-fold field function for geometry and color individually, which is defined as:
%---------------------------Equation-----------------------
\begin{equation}
     (\sigma, \mathbf{ H})=f_{\Theta_1}(x,y,z, t),
     \label{nerfsigma}
\end{equation}
\begin{equation}
    \mathbf{ c}=f_{\Theta_2}(\mathbf{H},\phi,\psi).
     \label{nerfcolor}
\end{equation}
%-----------------------------------------------------------
Similarly, $\Theta_1, \Theta_2$ denotes the trainable model parameters.

\subsection{T-Code for Dynamic Scenes}
\subsubsection{T-Code formulation}
As aforementioned, there remain three major difficulties for efficient dynamic view synthesis:  The first  is to represent temporal information in an efficient way without confusing across time interval. The second  is to cope with structural difference in the different domains. The third  is to handle differently structured observations in varied data. Targeting at these difficulties, we extend the multi-resolution hash encoding into the time domain. 

Denoting the encoded coordinates as $p \in \mathbf{R}^{d}$, multi-resolution hash encoding contains three steps: The first step is to get local features by locating the $2^d$ nearest grid points and getting hashed index according to the resolution. The second  is to fuse the features of nearest grid points by $d$-linear interpolation, mapping the continuous queried coordinates  into discrete, piece-wise linear feature vectors towards decoupled  fast training.  The third is to concatenate all features at multi resolutions for global perception. 

Let $N_{min}$ denote  the coarsest  resolution, \emph{i.e.} the resolution with the least grid points in one dimension, and $N_{max}$ denote the finest resolution with the most grid points in one dimension if available. 
 Resolutions at intermediate levels are obtained by scaling identically  between the two  neighboring levels if available.   For a spatial point $ \mathbf{x} =(x_1,x_2, \dots x_d) \in \mathbf{R}^d$ in $d$-dimensional  grids, we first find the nearest  grid points $p^i=(p^i_1,p^i_2,\dots p^i_d) \in \mathbf{R}^d,  i\in \{1,\dots 2^d\}$. Since features are stored in the discrete $p^{i}$ grid points,  coordinates  in every dimension are converted into integers, \emph{i.e.} $p^i_k\in \mathbf{Z},  k \in \{1,2,\dots d\}$. We then compute the feature indices $h_d(p^i)$  to access the features stored in the $p^i$ grids via:
% hash function
%---------------------------Equation-----------------------
\begin{equation}
    h_d(p)= (\bigoplus_{k=1}^{d} p_k\pi_k  ) \quad \mathbf{mod} \ \  T,
    \label{hash function}
\end{equation}
%-----------------------------------------------------------
where $\bigoplus$ denotes bit-wise XOR operation for integers, $p_k$ is the coordinate at the $k^{th}$ dimension,  $\pi_k$ is a large prime number,   $T$ is the length of the hash table,   and $\mathbf{mod}$ denotes modulo operation. This function maps $p^i\in \mathbf{R}^d$ to an integer index. Note that equation~\ref{hash function} holds for coordinates with arbitrary dimensions, and the symmetric XOR operation decorrelates  features across different  axes  \cite{ingp}.  Finally, the $d$-linear interpolation is performed to fuse the features of neighboring $2^d$  points. Let $h_d({P})$ denote the set of hashed indices of neighboring points, $[\;]$ denote the indexing operation, $interp()$ denote the $d$-linear interpolation, $\Psi_l$ denote the hash table corresponding to the $l^{th}$ resolution, the interpolated feature vector at each resolution is obtained by:
%---------------------------Equation-----------------------
\begin{equation}
\Psi_l(\mathbf{x})= interp(\Psi_l[h_d(P)]),
\label{indexing}
\end{equation}
%-----------------------------------------------------------
For integrating features at different resolutions, different levels of features are concatenated. Let $l \in \{1,\dots L\}$ denote the different levels, $\mathbin\Vert $ denotes vector concatenation, the encoded features $\Psi(\mathbf{x})$ are obtained through:
%---------------------------Equation-----------------------
\begin{equation}
    \Psi(\mathbf{x}) = \mathbin\Vert_{l=1}^{L} \Psi_l(\mathbf{x}),
    \label{concat}
\end{equation}
%-----------------------------------------------------------
where $\Psi_l(\mathbf{x}) \in \mathbf{R}^F$ with $F$ as the  predefined value.   The hash collisions are compensated by the downstream MLP.

The multi-resolution hash table structure restricts the number of parameters by predefined size $T$ and resolutions $L$.  The structure also enables parallel update of parameters, since all the queried features are obtained by similar interpolation within a subset of the hash table.  By simply applying equation~\ref{hash function} to equation~\ref{concat} in case of $d=1$, we yield  a simple yet effective representation for temporal information as:
%---------------------------Equation-----------------------
\begin{equation}
\Gamma_l(t)= interp(\Gamma_l[h_1((t))]),
\label{indexing_t}
\end{equation}
\begin{equation}
    \Gamma(t) = \mathbin\Vert_{l=1}^{L} \Gamma_l(t),
    \label{concat_T}
\end{equation}
%-----------------------------------------------------------
where $t\in \mathbf{R}^+$ is a scalar time stamp,  $\Gamma_l(t)$ denotes features at different resolutions and $\Gamma(t)$ is the entire   representation.

The proposed representation is straight-forward, and we name it "T-Code" due to the single time dimension encoded. 
This multi-level  encoding fuses temporal information in a compact and decoupled manner. T-Code  solves the three major difficulties for efficient dynamic view synthesis with three-fold advantages.

The first is the  strong  representational capacity for time dimension by fewer parameters in decoupled structure. Leveraging the independence of features in different time stamps, T-Code delivers a compact representation of individual time interval. Features in a time stamp are stored into a compact latent code of dimension $L\times F$, which is highly efficient and avoids aliasing  across large intervals. Moreover, our T-Code with a shared hash table structure is much more compact than the 1024-dim latent code for every frame in DyNeRF  \cite{Dynerf}, and the compactness enables fast training and inference.

The second  is the explicit decoupling from temporal to spatial features.  T-Code only parameterizes temporal information, and the explicit decomposition enables disentangled representation for time domain  and spatial domain individually. Moreover, since each time stamp in a dynamic scene could be considered as a static one, the dynamic representation is supposed to be self-contained at every frame. Fortunately, T-Code naturally matches this requirement, since it only focuses on temporal features, leaving the spatial ones to be represented by other modules. Thanks to the decomposition and interpolation, the temporal representation to be updated at every frame is reduced to $2^d$ tiny $L\times F$  vectors, and the property enables decoupled training at high speed. 
\label{sec: time and spatial domain_}

The third is the flexibility of model structure. Benefiting from the light parameter load and the strong representational capacity, T-Code is able to be combined with different modules to apply for different scenarios, which can achieve top training speed or performance improvement at minimum cost. The combined module could be specially chosen according to different data types. When combined with simple spatial hash encoding, T-Code is capable of achieving fast synthesis under multi-camera setting. When combined with deformation networks and hash grids, T-Code can boost existing representations under mono-camera setting.

\subsection{HybridNGP: Decoupled Representation for Multi-camera Dynamic View Synthesis}

%--------------------------------------------- Figure ------------------------------------
\begin{figure*}[!t]
\centering
\includegraphics[width=0.85\textwidth]{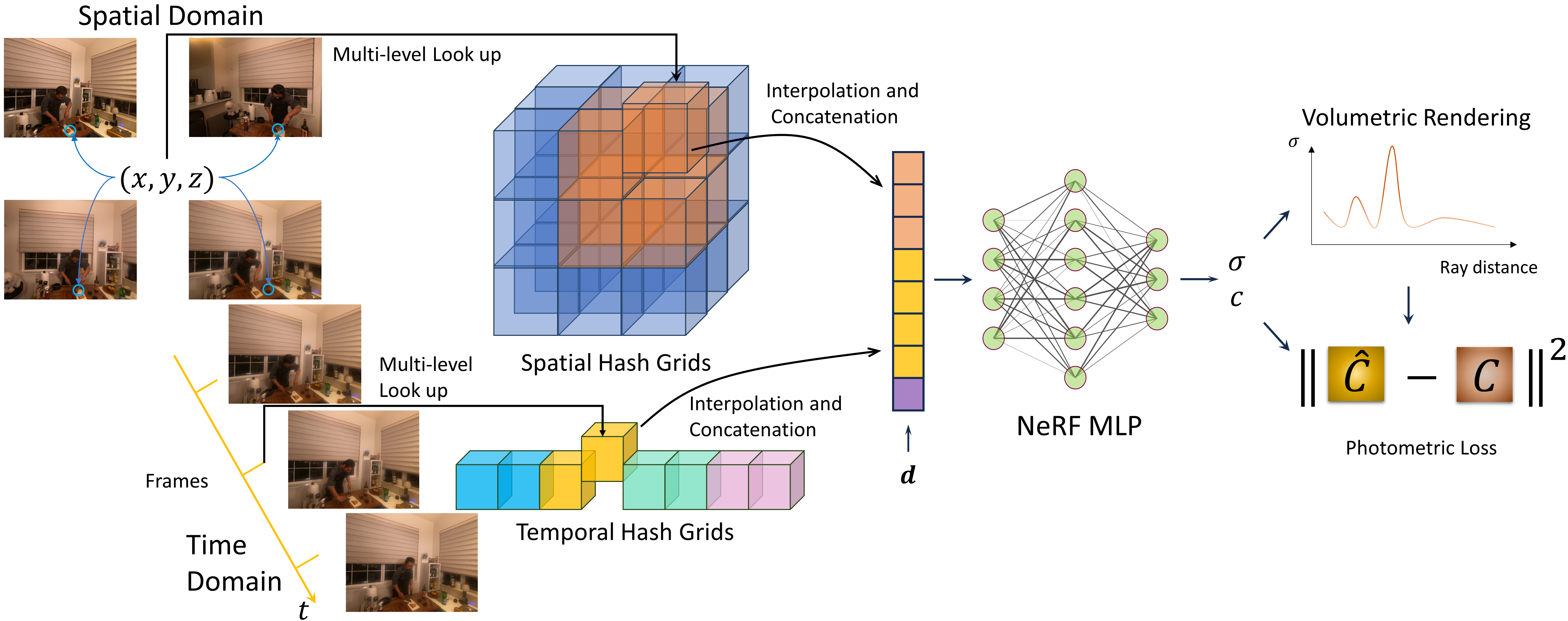} % Reduce the figure size so that it is slightly narrower than the column.
\caption{Outline of our HybridNGP model. HybridNGP consists of the T-Code and spatial hash grids.  
%The features are stored in spatial grids and T-Code independently.
A queried 4D point $(x,y,z,t)$ is splitted into the spatial $(x,y,z)$ and temporal $t$ queries. The spatial features are encoded by the spatial hash grids, while the temporal features are encoded by the T-Code. The spatial and temporal features are concatenated and then fed into the NeRF MLP for regressing $(\sigma, \mathbf{c})$. The output of NeRF MLPs are used for volume rendering described in equation~\ref{vren_discrete}. L2 photometric loss is used to optimize the model. The NeRF MLP adopts a 2-fold structure for $\sigma$ and $\mathbf{c}$ individually, as described in equation \ref{our_dnerf_sigma} and \ref{nerfcolor} respectively. $\mathbf{d}$ denotes encoded directions.  }
\label{fig:mainmodel}
\end{figure*}
%--------------------------------------------------------------------------------

\subsubsection{Model formulation.}
Our T-Code is compact, decoupled and representational for time features. It needs to be combined with other spatial representations for dynamic view synthesis. As shown in figure~\ref{fig:mainmodel}, we combine  T-Code with spatial hash grids to form our hybrid neural graphics primitives(HybridNGP) for multi-camera scenario.

As pointed out by   \cite{rerf}, simply using $\mathbf{x}=(x,y,z,t)$ to obtain the confused 4D hash feature is not suitable for high fidelity dynamic view synthesis. Therefore, we  compose a hybrid feature by simply concatenating the individual T-Code and spatial hash features to  avoid confusion in different axes. The concatenated features are then fed into NeRF MLP for volume rendering, formulated as: 
%---------------------------Equation-----------------------
\begin{equation}
     (\sigma,\mathbf{H})=f_{\theta_1}( \Psi(\mathbf{x}) \mathbin\Vert   \Gamma(t) ),
     \label{our_dnerf_sigma}
\end{equation}
%-------------------------------------------------
where $\Psi (\mathbf{x})$ denotes 3D spatial hash encoding in equation~\ref{concat}, $f_{\mathbf{\theta_1}}$ denotes a tiny MLP with trainable parameter $\mathbf{\theta_1}$. The NeRF MLP $f_{\mathbf{\theta_1}}$ fuses features across different domains and obtains the volume density $\sigma$. Colors are regressed by another tiny MLP defined in   equation~\ref{nerfcolor}. All parameters including  latent vector representations for different domains are expected to be optimized via observations from different directions in the multi-camera scenario. %Experiments prove that the HybridNGP with simple representation achieves highly desirable reconstruction quality, as well as minimum computational cost under multi-camera setting. 

\subsubsection{Rendering and optimization.}
Given the sampled optical attributes on the ray in equation \ref{cameraray}, we adopt  volume rendering by numerical integration to bridge the gap between 2D and 3D as:
%---------------------------Equation-----------------------
\begin{align}
    \hat{C}(r) &= \sum_{i=1}^N T_i\alpha_i \mathbf{c_i}= \sum_{i=1}^N T_i (1-e^{\sigma_i\delta_i}) \mathbf{c_i},
\label{vren_discrete}
\end{align}
%-------------------------------------------------
where $T_i=\exp(-\sum_{j=1}^{i-1} \sigma_j\delta_j)$ denotes the transmittance, $\delta_i$ denotes the interval between neighboring samples, $N$ denotes the number of samples along the ray. Upon the rendered pixels, the model are supervised and optimized using L2 photometric loss as: 
%---------------------------Equation-----------------------
\begin{equation}
    L = \frac{1}{|\mathbf{R}|} \sum_{r \in \mathbf{R}} \|C(r) - \hat{C}(r)\|_2^2 
    \label{eq:optim}
\end{equation}
%--------------------------------------------------
where $\mathbf{R}$ denotes the set of the camera rays. Besides L2 loss, we apply three regularization terms for our HybridNGP. The first term $L_d$ is the distortion loss  in  Mip-NeRF 360  \cite{mipnerf360}, defined as:
\begin{align}L_{d} &= \sum_{i,j} w_i w_j  \left\lvert \frac{s_i + s_{i+1}}{2} - \frac{s_j + s_{j+1}}{2} \right\rvert \notag  \\
&+ \frac{1}{3} \sum_i w_i^2 (s_{i+1} - s_i)
\end{align}
\noindent where $s_i$  denotes the normalized ray distance along the camera ray, $w_i=T_i \alpha_i$ denotes the weight of the $i^{th}$ sample. 
The second term $L_o$ is the entropy on the opacity   by   \cite{ngppl}, defined as:  
%---------------------------Equation-----------------------
\begin{equation}\label{opac}%\label{opac_loss}
  L_{o} = -o\log(o), \quad o = \sum_{j=1}^{N}T_j\alpha_j.
\end{equation}
%--------------------------------------------------
The third regularization term $L_s$ is the  binary entropy  of $\sigma$ to remove floaters and alleviate foggy effects in dynamic scenes, defined as:
%---------------------------Equation-----------------------
\begin{align}
L_{s} &= -\rho\log(\rho) - (1-\rho)\log(1-\rho) \label{entropy},
\end{align}
%--------------------------------------------------
where $\rho$  is  clipped and normalized $\sigma$ from $(0,10)$ to $(0,1)$.
 Based on the configuration in \cite{ngppl}, we use  $L_o$ with weight 0.005 to remove foggy effects, and  $L_d$ with weight 0.0005 for allowing transient objects in dynamic scenes. 
To ensure numerical stability and obtain better results,  we set the  weights of $L_s$ as 0.005.

\subsection{T-Code for Monocular Dynamic View Synthesis}
T-Code is not only suitable for multi-camera dynamic view synthesis, but also applicable to monocular settings.
 However, 
 The monocular video
 synthesis requires building cross-time relationship, which is
 not considered in the simple concatenated hash encoder and
 MLP decoder design in HybridNGP.
   We thus introduce deformation-based NGP with T-Code (DNGP-T) based on D-NeRF  \cite{Dnerf} and Torch-NGP  \cite{torch-ngp}. To cope with monocular scenarios, DNGP-T utilizes a tiny deformation network like  \cite{Dnerf}, and applies T-Code to  cope with  appearance changes.  DNGP-T is defined as:
%---------------------------Equation-----------------------
\begin{align}
 (\sigma,\mathbf{H}) &=f_{\theta_1}( \Psi( g_d(\mathbf{\eta,\tau}) ),\eta) \label{dngp_sigma} \\
    \mathbf{c} & =f_{\theta_2}(\mathbf{H},\mathbf{\tau},  \Gamma(t),\mathbf{d}) \label{dngnp_c}
\end{align}
%--------------------------------------------------
where $\eta,\mathbf{\tau}$ are Fourier-encoded space coordinates $\mathbf{x}\in \mathbf{R}^3$ and time stamp $t\in \mathbf{R}^+$, $g_d(\mathbf{\eta,\tau})$ denotes the deformation MLP, $\mathbf{d}$ denotes the encoded directions. % by sphere-harmonics \cite{nerfeccv, ingp}.
The primary difference between HybridNGP and DNGP-T is that DNGP-T applies deformation network to encode the motion and geometry, and the T-Code in DNGP-T models appearance differences across frames. The model is optimized by a simple L2 loss as shown in equation \ref{eq:optim} without other regularization. \label{sec: dngp_t description}

\section{Experiments}

%We conduct experiments to prove the viability of our T-Code. We first conduct experiments on HybridNGP for multi-camera setting, we then combine T-Code to compose DNGP-T for mono-camera setting.

\subsection{HybridNGP for Multi-camera Dynamic View Synthesis}
\subsubsection{Datasets.}

Currently, there are only DyNeRF dataset  \cite{Dynerf} providing high quality multi-camera videos for dynamic view synthesis, and most of the current state-of-the-art methods are evaluated only on DyNeRF dataset for multi-camera scenario. We thus only test our HybridNGP on the DyNeRF dataset. DyNeRF dataset contains 6 individual cooking scenes with varied illumination, food topological changes and transient flame. Each scene contains at least 300 frames of video in 30 FPS, with up to 21 synchronized camera views at the resolution of 2028 $\times$ 2704.

\subsubsection{Implementation details.}
Our implemetation is based on HexPlane  \cite{HexPlane_}, Instant-NGP  \cite{ingp} and NGP-pl   \cite{ngppl}. All the metrics are measured on a single RTX 3090 GPU if not otherwise specified. Similar to the previous work  \cite{HexPlane_,Dynerf}, we select the video with index $0$ for evaluation. Following HexPlane  \cite{HexPlane_}, we conduct experiments  on 5 synchronized scenes except for the \emph{coffee martini} scene.  To fit the videos in memory, we downsample the videos to the resolution of $\mathbf{512\times384}$.  Metrics produced in higher resolutions may be  slightly downgraded, but there is no substantial difference. 
The spatial resolutions $N_{min}, N_{max}$ are set as $16$ and $2048$ times the scale of the scene. The size of the hash tables, \emph{i.e.} $T$ is set as $2^{19}$ for spatial feature, and $2^9$ for T-Code to trade off between feature sharing and model size.  $L, F$ for the spatial encodings are set as $16, 2$.
The sigma MLP defined in equation \ref{our_dnerf_sigma} contains 2 hidden layers. The length of the latent feature $\mathbf{H}$ in equation \ref{our_dnerf_sigma} is set as 48. The RGB MLP defined in equation~\ref{nerfcolor} contains only one hidden layer. The widths of hidden layers are all set as 64 neurons if not otherwise specified.  
To ensure numerical stability and yield best quality, the model is optimized by Adam \cite{adamoptim}  with $\epsilon=10^{-15}$,  and the  weight decay of the MLP and hash table parameters are chosen as $1\times 10^{-7}$  and  $5\times 10^{-8}$ respectively.  To get better  generalization across all frames, we apply the distortion loss $L_d$  \cite{mipnerf360} only after  step $18k$. The learning rates are set as 0.001 and scheduled by cosine function. The random seeds are set as  1337.

 \label{sec:Implementation details}
 
 To accelerate the training, we adopt ray marching strategy by maintaining an occupancy grid in our HybridNGP similar to  \cite{ingp}. Since all the information of the scene are stored in the hash tables and MLPs,  the occupancy grid only serves as a cache and may be deleted to save storage.  
 To prevent collapse,  we  choose the first 4096 iterations as the warming up stage. During the warming up stage, we sample the 3D points across the whole 3D scene, otherwise we sample points according to the occupancy grid. To simplify the model, the occupancy grid is shared across all frames and is updated after every 16 iterations.
 We adopt a simplified hierarchical sampling strategy from those of  \cite{Dynerf,HexPlane_}.  The sampling probabilities are only  computed according to Geman-McLure robust function of the difference from the global mean, shown as:
 \begin{equation}
     W_t(r)= \frac{1}{3} \|  \nu( (C_t(r) - \Bar{C}(r); \gamma))  \|_1
     \label{sampling strategy}
 \end{equation}
 \noindent where $\nu(x; \gamma)= \frac{x^2}{x^2+\gamma^2}$ is the element-wise robust  cost function,  $W_t(r)$ is the relative weight of the pixel, $C_t(r)$ is the pixel in image  at frame $t$, $\Bar{C}(r)$ is the cached global average across all frames of the pixel.  $\gamma$  is a scaling factor predefined to 0.02. 
 The ratio of randomly sampled pixels to all pixels are increased from initial $\frac{1}{2}$ to $\frac{3}{4}$ and $\frac{7}{8}$ at the $36k, 54k$ step respectively.

\subsubsection{Quantitative comparison.}
We report three quantitative metrics: PSNR, DSSIM, and LPIPS-AlexNet  \cite{LPIPS} for synthesis quality evaluation. Table~\ref{Dynerf_comparison} shows the comparison between the state-of-the-art methods and our HybridNGP.  The evaluation results of HexPlane are reproduced directly from their official  implementation. The others are copied from the previous publications if available. HexPlane-S is a cloned HexPlane model with only difference on ray batch size for acceleration.
\label{sec:quantitative comp}

%--------------------------------------------- Table ------------------------------------
\begin{table*}[ht]
    \centering
    \resizebox{\textwidth}{!}{%
    
    \begin{tabular}{llllllll}
        \Xhline{3\arrayrulewidth}
        \textbf{Method}  & \textbf{Rays} & \textbf{Steps} & \textbf{PSNR $\uparrow$} & \textbf{DSSIM $\downarrow$} & \textbf{LPIPS $\downarrow$} & \textbf{Time $\downarrow$} & \textbf{Storage $\downarrow$} \\ \hline
    DyNeRF  \cite{Dynerf}           & 4096                    & 650k           & 29.58        & 0.020           & 0.099          & $\sim$ weeks       & 28MB                     \\
    LLFF  \cite{llff}                                                                                         & -                              & 200k                            & 23.24                                     & 0.0762                                        & 0.2346                                       & -                                               & -                                              \\
    
    NeRF-T  \cite{Dynerf}           & -                       & -              & 28.45        & 0.023           & 0.10          & -                 & -                        \\ 
    Interpolation-NN  \cite{park2023temporal} & 6144                    & 600k           & \textbf{32.15}        &      -         & 0.0731        & $\sim$ days        & \textbf{20MB}                     \\ 
    HexPlane-S       & 1024                    & 100k           & 31.33      & 0.0159         & 0.1107       & 56min             & 327MB                    \\
    HexPlane  \cite{HexPlane_}                  & 4096                    & 100k           & 32.12      & 0.0127        & 0.0839       & 96 min             & 327MB                    \\ 
    K-Planes  \cite{kplanes}     & 4096                    & 90k            &   31.63            &          -       &     -           & 110 min            & 200MB                    \\ 
    StreamRF  \cite{streamrf}         & 5k                      & 128k           & 28.26         &  -              &  -              & 75 min             & 5000MB                   \\ 
    Naive4DNGP      & 512                     &  90k              & 23.67         & 0.099           & 0.48           & 60min             &  43MB                       \\ 
    Ours-45k       & 512                     & 45k            & 31.41        & 0.0131    
        & 0.0655       & \textbf{20min}             & 43MB             
                \\ Ours-90k 
                &           512                   & 90k             & 31.64        & \textbf{0.0126}        & \textbf{0.0610}      & 40min             & 43MB                     \\ 
                     \Xhline{3\arrayrulewidth}
    \end{tabular}      
    }           
    \caption{Quantitative comparison of the results produced by our HybridNGP and other methods on the DyNeRF dataset.  Benefiting from the feature decomposition by  T-Code,  HybridNGP yields desired reconstruction quality
     while  achieving the fastest training with much smaller checkpoint
      than previous fast methods. \emph{Rays} denotes the number of pixels 
      sampled in one batch.  \emph{Steps} denotes iterations in optimization.
       DyNeRF uses 8 V100 GPUs and Interpolation-NN uses 8 V100 or equally 4 A100 GPUs, and both of them provide no official code release. 
       NeRF-T has no official release, LLFF\cite{llff} is designed for static scenes and produces poor quality,  thus we only adopt their per-frame quality metrics from previous publication \cite{Dynerf}. 
       Due to the the limitation of our  platform for experiments, we are unable to run K-Planes,  and StreamRF, thus we are unable to report the missing metrics in their publications. }
    \label{Dynerf_comparison}
    \end{table*}
    %---------------------------------------------------------------------------------
    
Our HybridNGP  achieves the fastest training with high reconstruction quality. With the T-Code, our HybridNGP trained for 45k steps outperforms HexPlane-S with close performance to  HexPlane. After 90k steps, HybridNGP yields comparable metrics to HexPlane, while consuming much less GPU time. Moreover, to support the significance of the feature decomposition between  spatial domain and  time domain, we also report metrics from Naive4DNGP. The only difference between Naive4DNGP and HybridNGP is that we replace the decoupled hash encoding in HybridNGP shown in equation \ref{our_dnerf_sigma} with the naive hash grids for 4D space $(x,y,z,t)$ in equation \ref{hash function}, \ref{indexing} and \ref{concat}. It is obvious that the tangled 4D hash-encoding leads to a significant quality drop, showing the superiority of the decomposed feature.

Another advantage of  HybridNGP is the low storage consumption:  the tensor representations  \cite{HexPlane_, kplanes} yield state-of-the-art reconstruction quality at high speed, but their models are much larger. Although  applicable for incremental setting, SteamRF  \cite{streamrf} achieves fast training with quality loss and huge storage cost. With the compact T-Code,  HybridNGP consumes the lowest storage except for the slowest NN-based methods   \cite{park2023temporal,Dynerf}. 
This is because geometry patterns along $(x, y, z)$ axes remain
congruent even if the axes are transposed, while $(x, y, z, t)$
axes loses congruence in physics after any transpose operation related to t axis. So, it is more efficient to leverage the
property, and learn 4D features via separated and asymmetric encodings instead of symmetric ones like \cite{HexPlane_}.

\subsubsection{Qualitative analysis.}
\begin{figure*}[ht]
    \centering
    \begin{subfigure}[b]{0.24\textwidth}
        \centering
        \includegraphics[width=\textwidth]{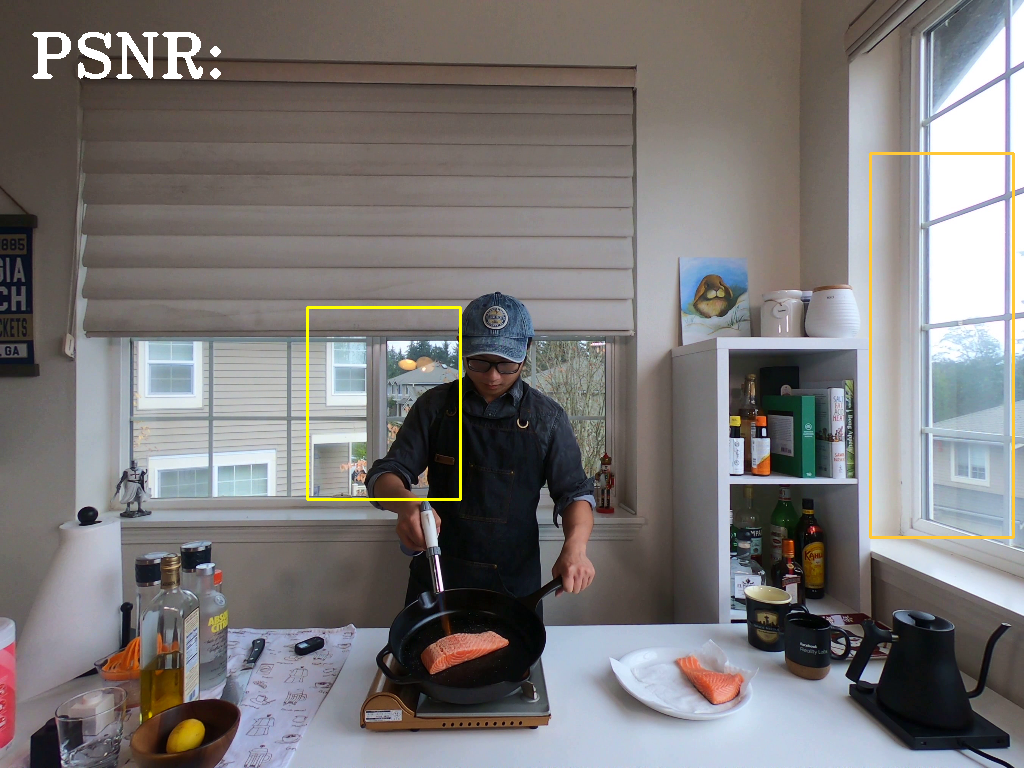}
        \caption{Ground truth.}
        \label{fig:gt}
    \end{subfigure}
    \begin{subfigure}[b]{0.24\textwidth}
        \centering
        \includegraphics[width=\textwidth]{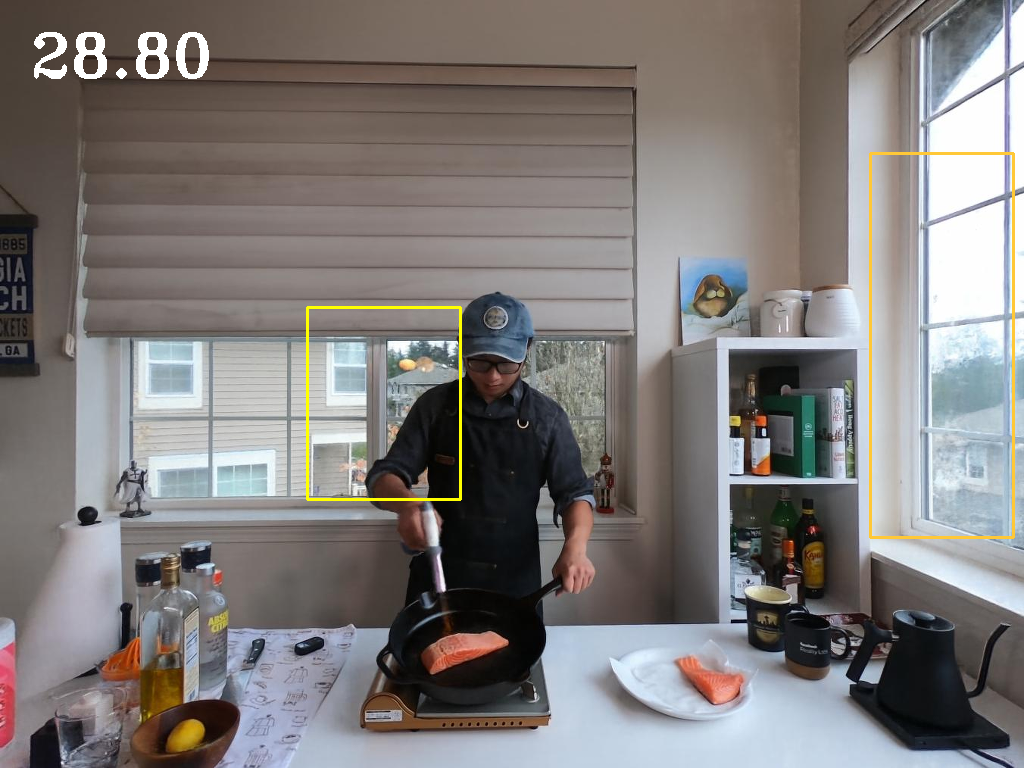}
        \caption{DyNeRF}
        \label{fig:DyNeRF}
    \end{subfigure}
    \begin{subfigure}[b]{0.24\textwidth}
        \centering
        \includegraphics[width=\textwidth]{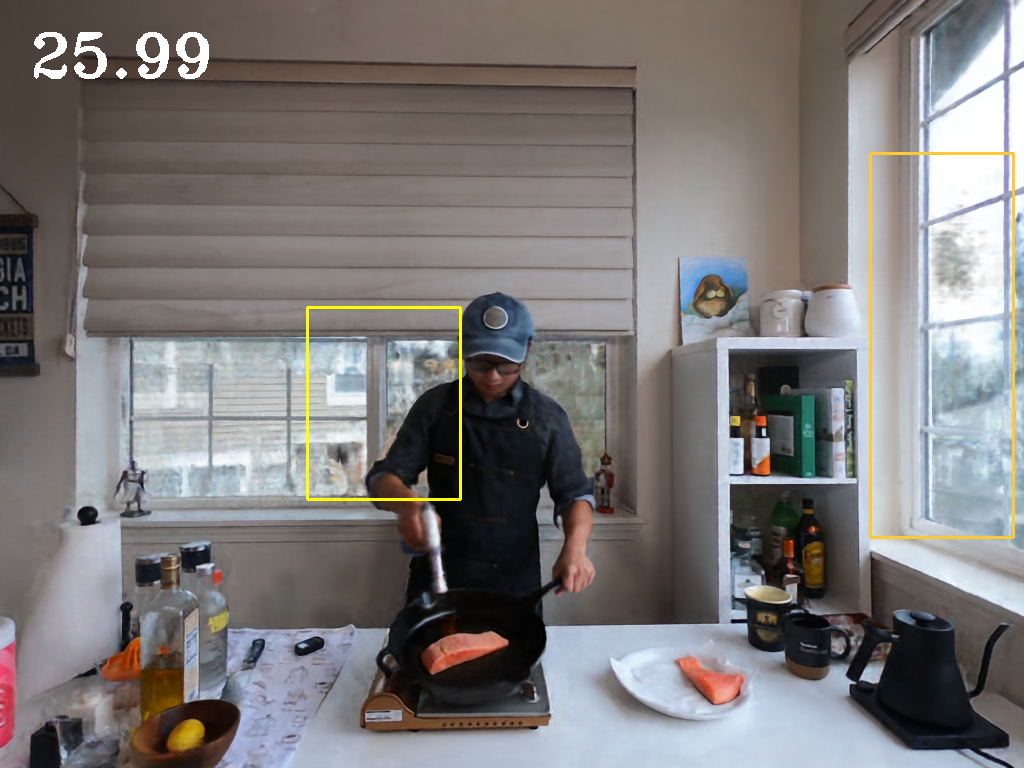}
        \caption{HexPlane-S}
        \label{fig:HexPlane-s}
    \end{subfigure}
    \begin{subfigure}[b]{0.24\textwidth}
    \centering
    \includegraphics[width=\textwidth]{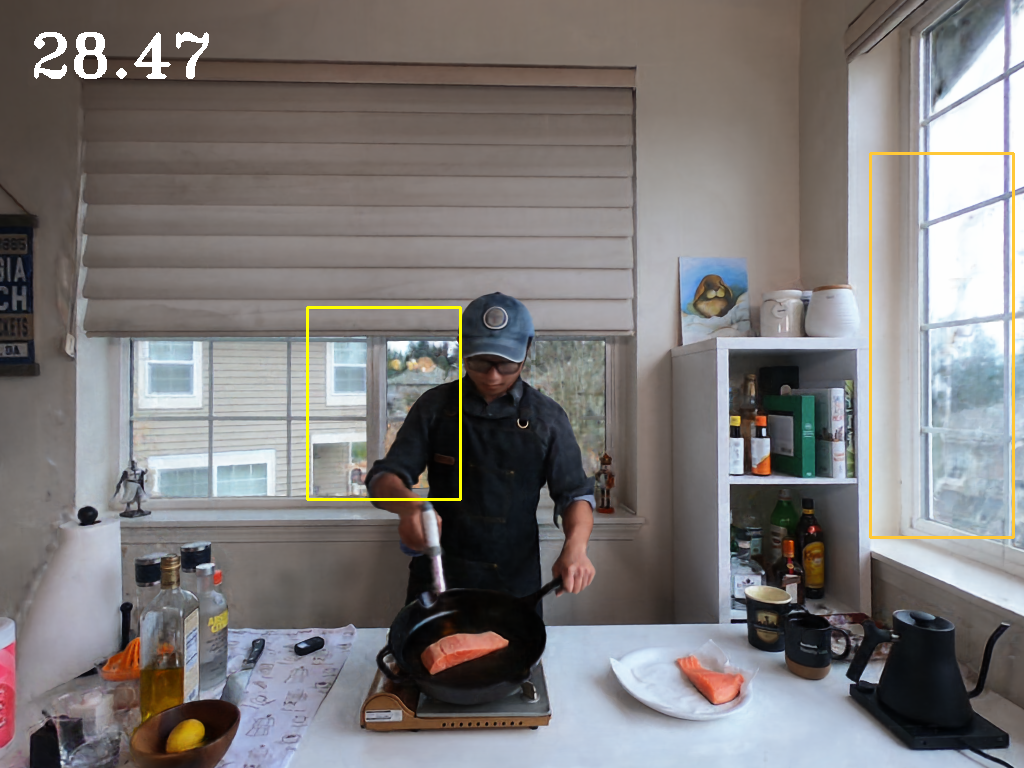}
    \caption{HexPlane}
    \label{fig:HexPlane}
    \end{subfigure}
    
    \begin{subfigure}[b]{0.24\textwidth}
    \centering
    \includegraphics[width=\textwidth]{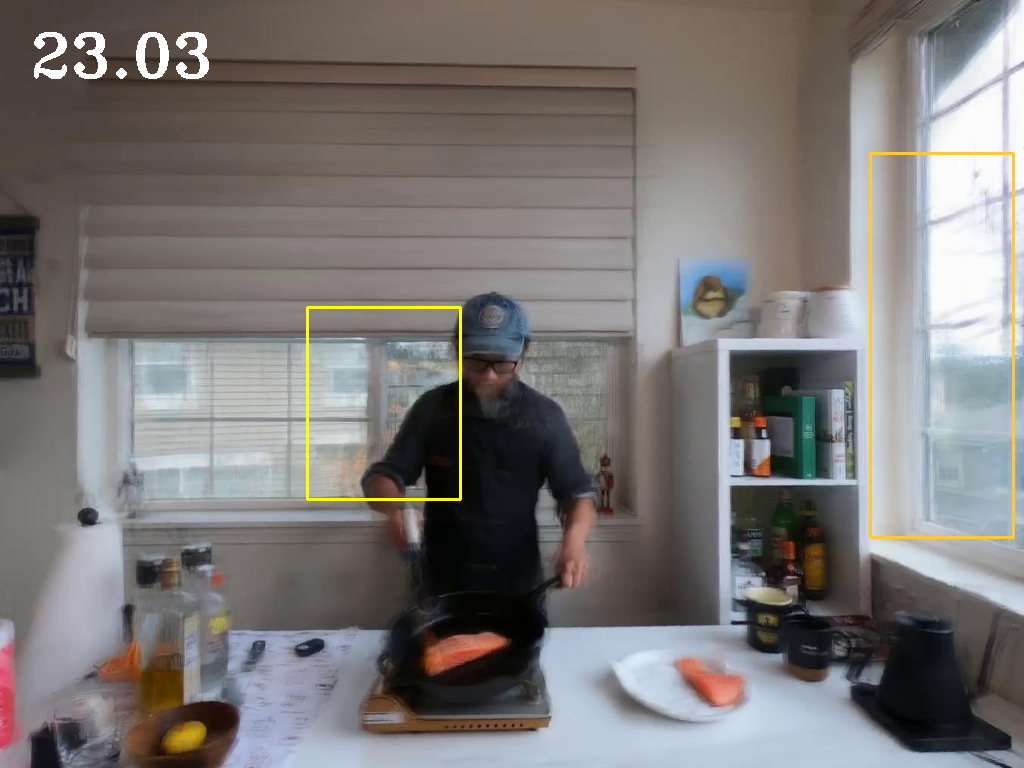}
    \caption{LLFF}
    \label{fig:LLFF}
    \end{subfigure}
    \begin{subfigure}[b]{0.24\textwidth}
    \centering
    \includegraphics[width=\textwidth]{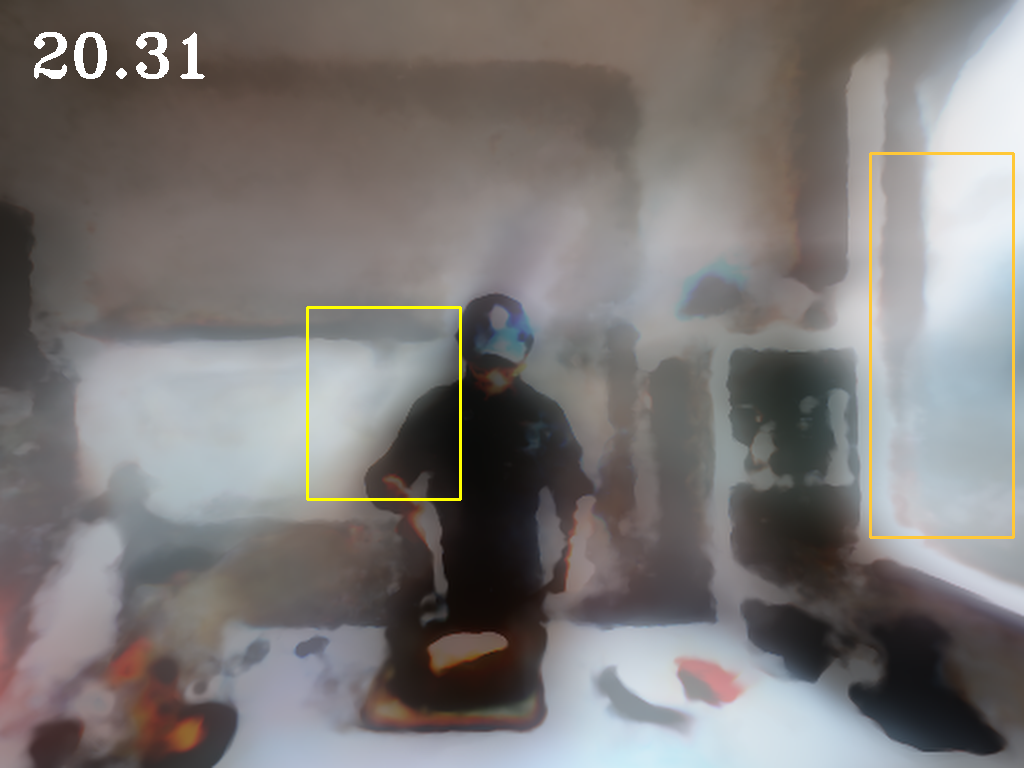}
    \caption{Naive4DNGP-90k}
    \label{fig:naive4d}
    \end{subfigure}
    \begin{subfigure}[b]{0.24\textwidth}
        \centering
        \includegraphics[width=\textwidth]{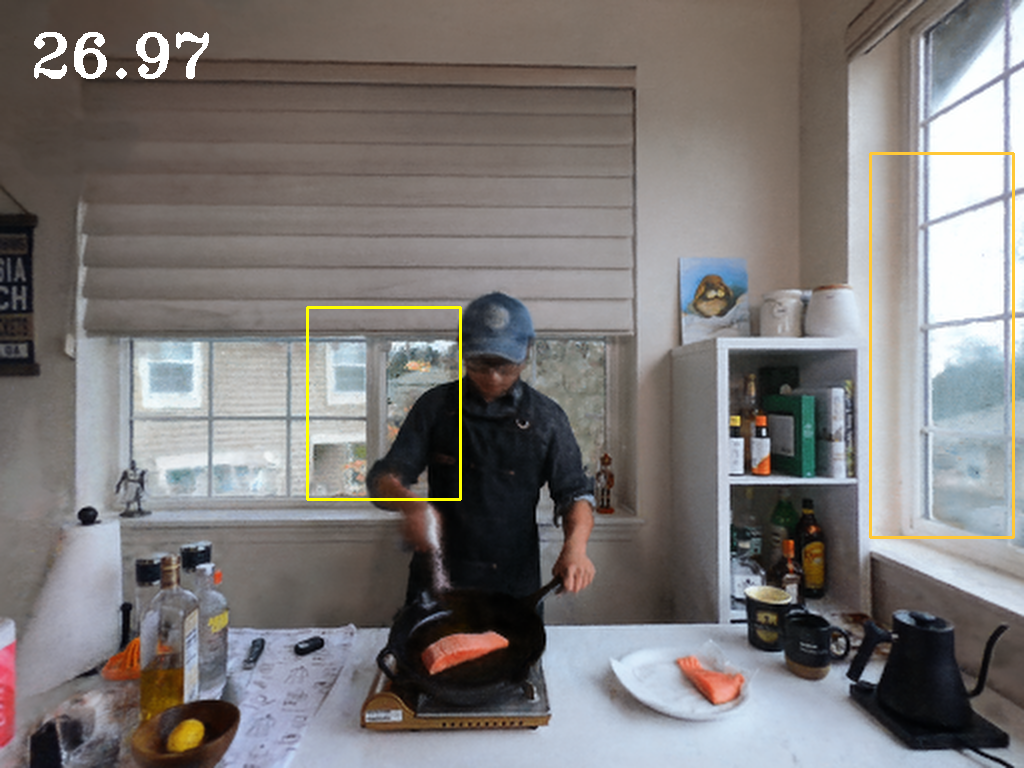}
        \caption{HybridNGP-45k}
        \label{fig:hybridngp-45k}
    \end{subfigure}
    \begin{subfigure}[b]{0.24\textwidth}
    \centering
    \includegraphics[width=\textwidth]{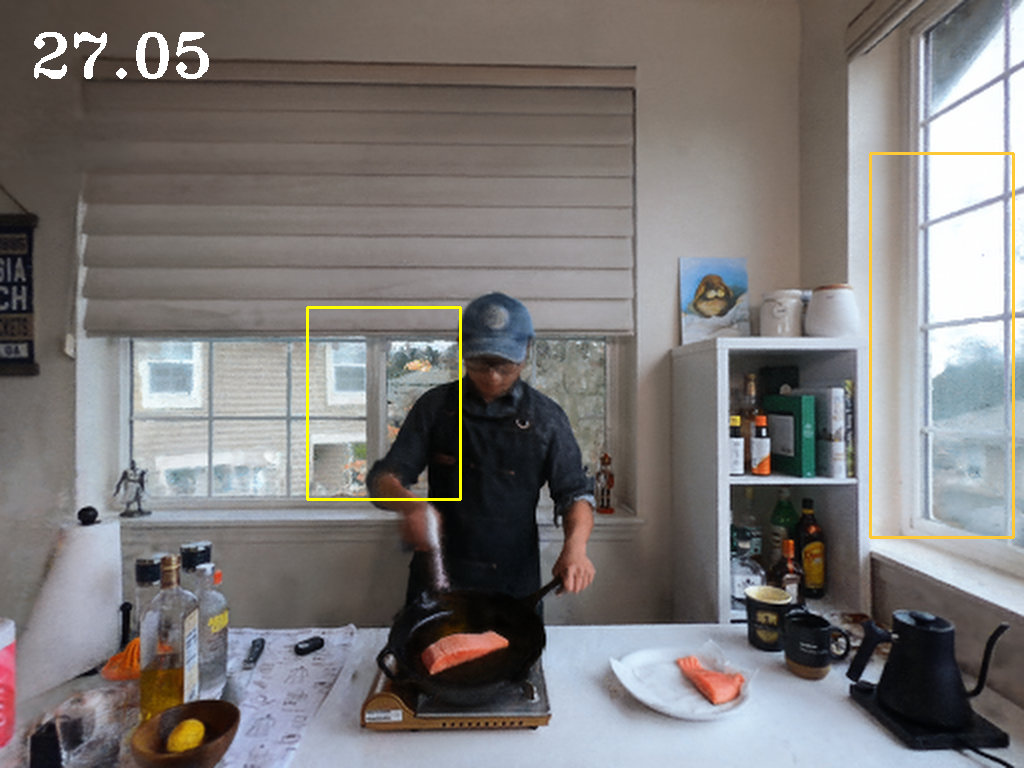}
    \caption{HybridNGP-90k}
    \label{fig:hybridngp-90k}
    \end{subfigure}
    
    \caption{Comparison of reconstruction quality on DyNeRF dataset. HybridNGP delivers result with more accurate local patterns (the reflections of light bulbs on  window and the outdoor scenes). Please note the PSNR values are  computed under the resolution of $1024\times768$ for fair comparison. }
    \label{fig:figure_comparison}
\end{figure*}

We provide the rendered images of HybridNGP, and compare them against other methods in figure \ref{fig:figure_comparison}. As shown in the figure, although consuming much less GPU time and storage,  HybridNGP delivers results at least comparable to HexPlane-S, which fail to recover local patterns precisely. The comparison between Naive4DNGP and HybridNGP also supports the representational capacity of  T-Code and the superiority of the decomposed representation, since learning separate representations removes the tangling in different axes. The mismatch between the perceptual quality and the quantitative metrics in table \ref{Dynerf_comparison} is caused by different resolutions, but there is no essential difference.

%--------------------------------------------- Figure ------------------------------------
%---------------------------------------------------------------------------------

\subsubsection{Ablation study on  T-Code structure.}

%--------------------------------------------- Figure ------------------------------------
\begin{figure}[!htbp]
    \centering
        \begin{subfigure}[b]{0.4\columnwidth}
        \centering
        \includegraphics[width=\textwidth]{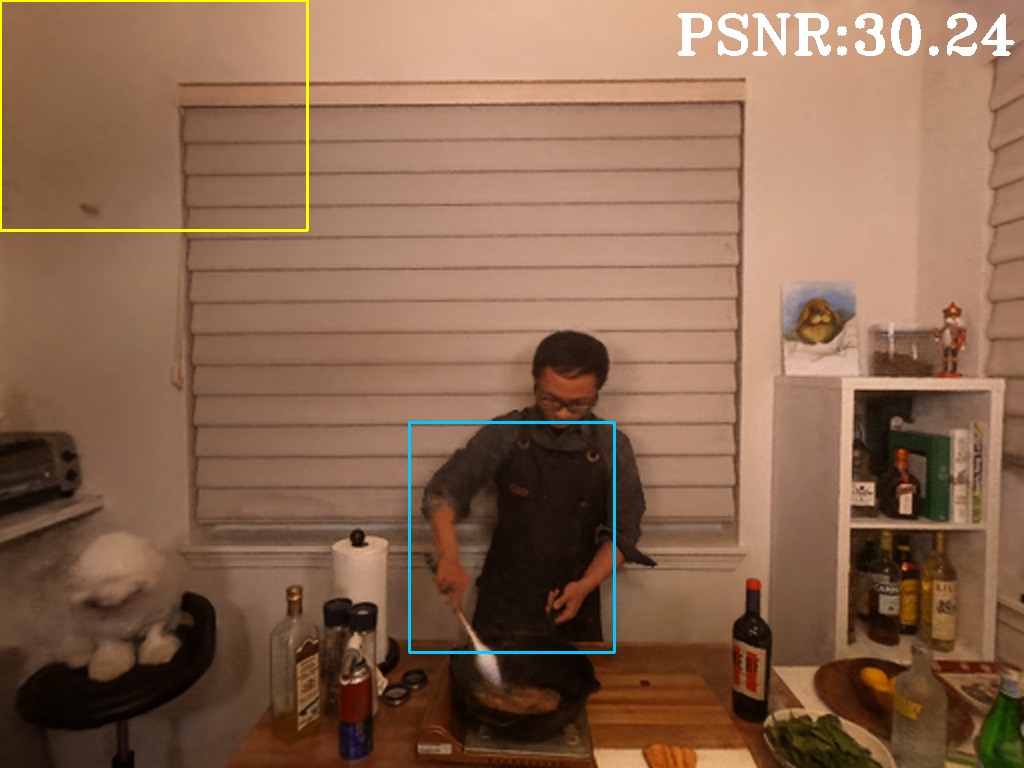}
        \caption{ $N_{min}=90$}
        \label{fig:nmin90}
    \end{subfigure}
    \begin{subfigure}[b]{0.4\columnwidth}
        \centering
        \includegraphics[width=\textwidth]{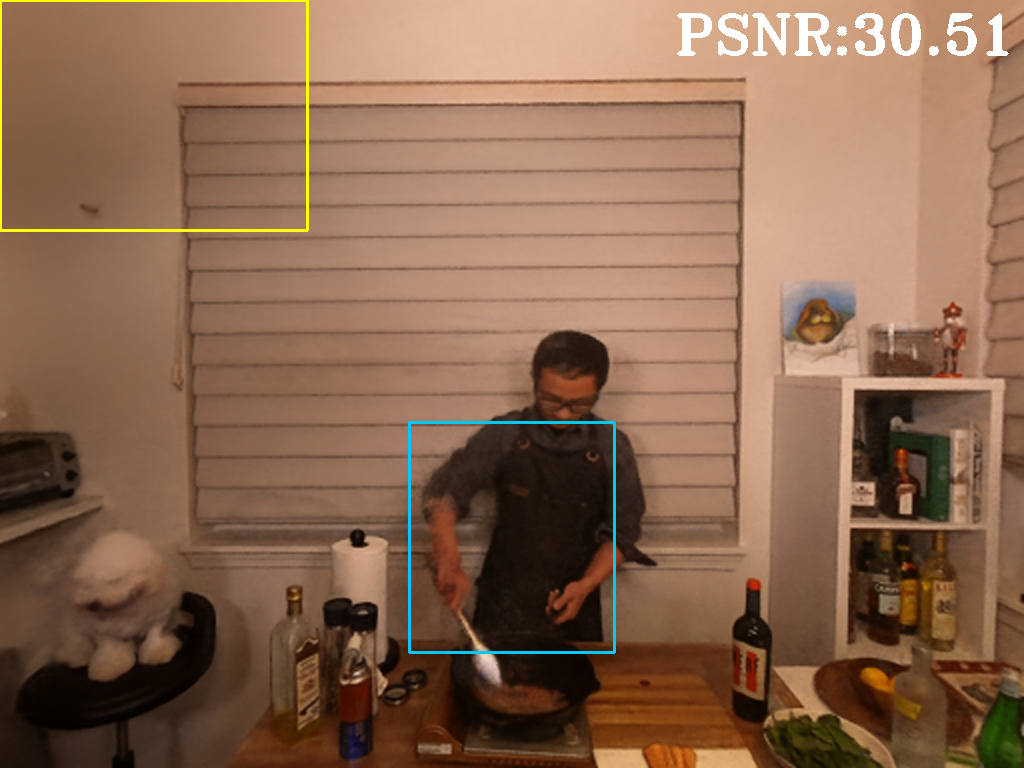}
        \caption{ $N_{min}=120$}
        \label{fig:nmin120}
    \end{subfigure}

    \begin{subfigure}[b]{0.4\columnwidth}
        \centering
        \includegraphics[width=\textwidth]{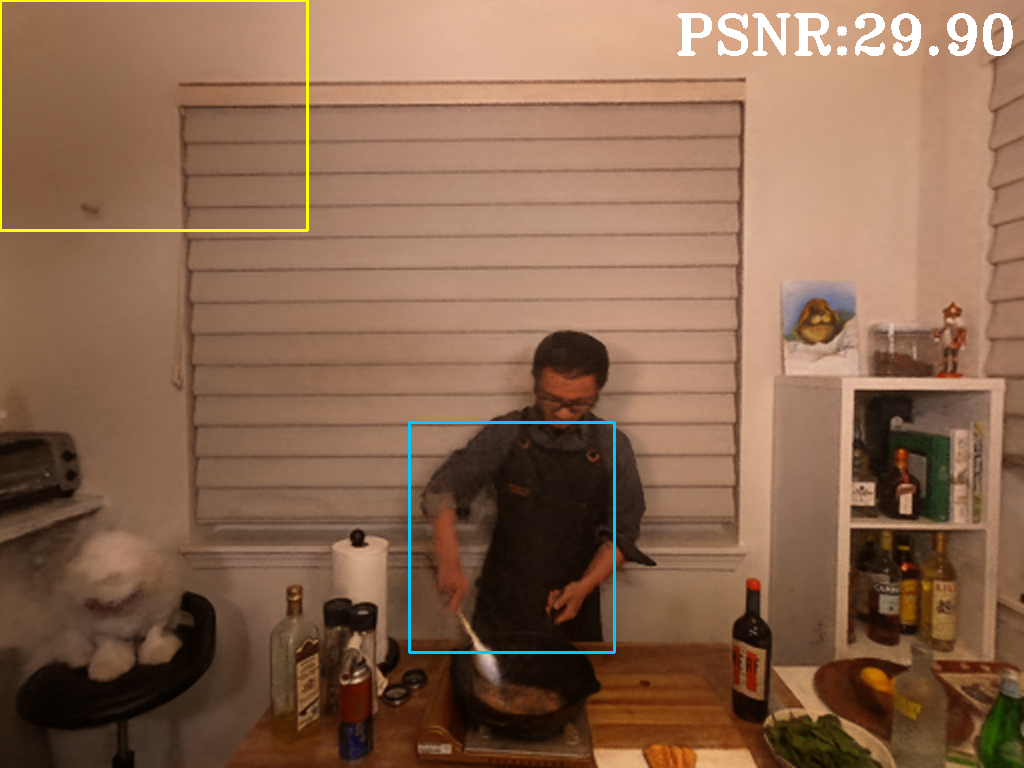}
        \caption{ $N_{min}=200$}
        \label{fig:nmin200}
    \end{subfigure}
    \begin{subfigure}[b]{0.4\columnwidth}
    \centering
    \includegraphics[width=\textwidth]{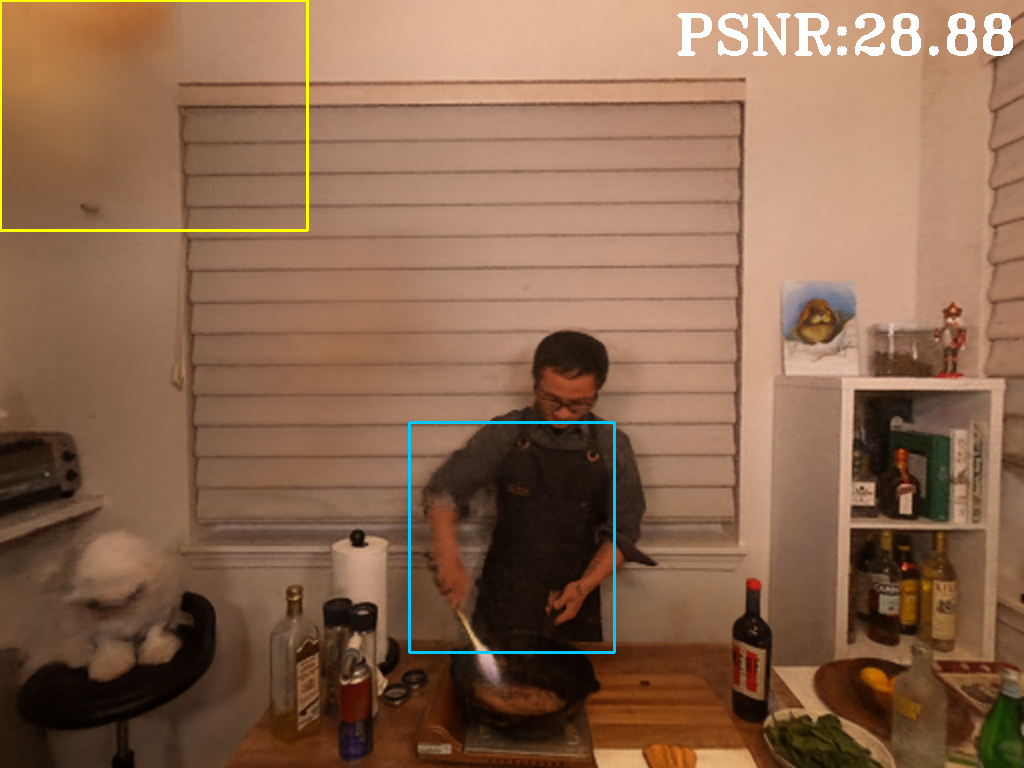}
    \caption{$N_{min}=300$}
    \label{fig:nmin300}
    \end{subfigure}
       \caption{Rendered results for a certain frame under different $N_{min}$. Benefiting from the interpolation, T-Codes with smaller $N_{min}$ generalizes better and avoids artifacts. }
       \label{fig:ablation}
\end{figure}
%--------------------------------------------- ------------------------------------

%--------------------------------------------- Figure ------------------------------------
\begin{figure}[htbp]
    \centering
    \begin{subfigure}[b]{0.24\columnwidth}
        \centering
        \includegraphics[width=\textwidth]{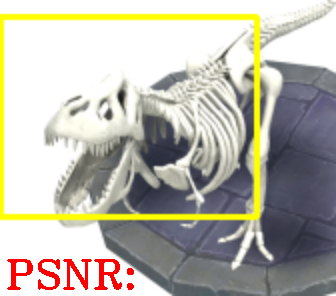}
        \caption{GT}
        \label{fig:trex_gt}
    \end{subfigure}
    \begin{subfigure}[b]{0.24\columnwidth}
    \centering
    \includegraphics[width=\textwidth]{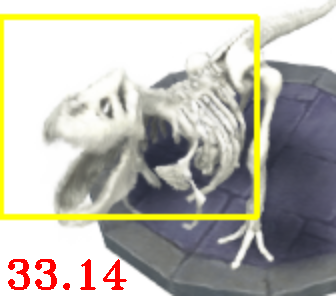}
    \caption{TineuVox}
    \label{fig:trex_tineu}
    \end{subfigure}
        \begin{subfigure}[b]{0.24\columnwidth}
    \centering
    \includegraphics[width=\textwidth]{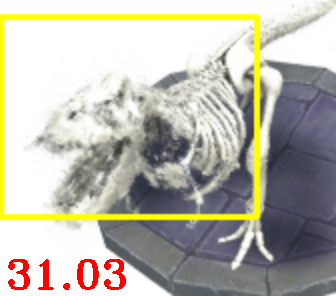}
    \caption{Torch-NGP}
    \label{fig:trex_torchngp}
    \end{subfigure}
    \begin{subfigure}[b]{0.24\columnwidth}
    \centering
    \includegraphics[width=\textwidth]{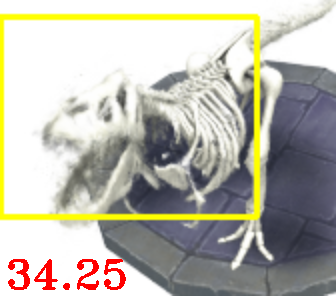}
    \caption{DNGP-T}
    \label{fig:trex_dngpt}
    \end{subfigure}
       \caption{ Rendered \emph{trex} scene in  D-NeRF dataset. DNGP-T recovers the skull and ribs better than the other methods. }
       \label{fig:trex_comp}
\end{figure}
%--------------------------------------------------------------------------------

We  provide more evidences to support the superiority of the T-Code representation with evaluation results listed in table \ref{tab:ablation}. Comparison between   the rendered image with different $N_{min}$ configurations is shown in figure \ref{fig:ablation}.  HybridNGP with a single-resolution T-Code outperforms the others, but the difference is subtle. Similar metrics reflect the robustness of the T-Code against structural difference.  However, the resolution in time axis is of significance. T-Code with highest resolution performs much worse than the others, indicating the interpolation in different stamps is beneficial for generalization. The similarity among $N_{min}= 90, 120$ and 200 presents the robustness of the T-Code against hyper parameters. The significance of interpolation is identical with  the conclusion of \cite{park2023temporal}.

%--------------------------------------------- Table ------------------------------------
\begin{table}[t]
    \centering
    \begin{tabular}{lccc}
    \Xhline{3\arrayrulewidth}
    \textbf{Architecture} & \textbf{PSNR $\uparrow$} & \textbf{DSSIM $\downarrow$} & \textbf{LPIPS $\downarrow$} \\ \hline
    $L=1,\quad F=40$             & \textbf{31.41}        & {0.0131}        & \textbf{0.0655}       \\ 
    $L=2,\quad F=20$                     & 30.99        & 0.0138        & 0.0687       \\ 
    $L=5,\quad F=8$                     & 31.21        & \textbf{0.0130}        & 0.0679       \\    
    
    $N_{min}=90 $                  & 31.17                                         & 0.0142                                                            & 0.0714                                                   \\
    $N_{min}=200$                    & 31.16                                          & 0.0133                                                              & 0.0695                                                 \\
    $N_{min}=300$                   & 30.74                                           & 0.0147                                                            & 0.0736                                                   \\
    \Xhline{3\arrayrulewidth}
    \end{tabular}
    \caption{Ablation study on different configurations of the  T-Code in HybridNGP. $L$ denotes the number of resolutions in  T-Code, and $F$ denotes the length of the  vector in each resolution. $L, F, N_{min}$ are set to 1, 40, 120 by default.  $N_{max}$ is set to 450 to remove the interpolation in  T-Code  of the 300 video frames. All metrics are computed at step 45k. }
    \label{tab:ablation}
\end{table}
%--------------------------------------------- -----------------------------------

\subsubsection{Ablation study on HybridNGP structures}

%-----------------------table-----------------
%---------------------------------------

%-----------------------table-----------------
\begin{table}[htbp]
    \begin{tabular}{lcccc}
    \Xhline{3\arrayrulewidth}    \textbf{Layout}&\textbf{PSNR$\uparrow$}& \textbf{DSSIM$\downarrow$}&\textbf{LPIPS$\downarrow$}&\textbf{Time $\downarrow$} \\ \hline
    $(12,2,1,40)$                                                                                     & \textbf{31.41}                                     & 0.0131                                        & 0.0655                                       & \textbf{20min}                                       \\
    $(12,2,1,32)$                                                                                      & 30.40                                     & 0.0165                                        & 0.0888                                       & \textbf{20min}                                       \\
    $(12,2,1,48)$                                                                                      & 30.72                                     & 0.0154                                        & 0.0826                                       & \textbf{20min}                                       \\
    $(12,2,1,64)$                                                                                      & 30.54                                     & 0.0171                                        & 0.1108                                       & 23min                                       \\
    $(12,4,1,64)$                                                                                      & 31.04                                     & 0.0122                                        & 0.0606                                       & 24min                                       \\
    $(12,4,1,80)$                                                                                                    &   30.36                                   &                 0.0138                              &                 0.0657                                  &      26min  \\
    $(12,4,1,80)^\dag$                                                                                                  &   \textbf{31.41}                                  &                \textbf{0.0116}                              &                 \textbf{0.0567}                                  &      26min                          \\ \Xhline{3\arrayrulewidth}
    \end{tabular}
    \caption{Comparison on different HybridNGP structures. The 4 numbers in  the \emph{Layout} tuple  denotes $L, F$ for spatial hash encoding and $L, F$ for T-Code respectively.  Metrics are recorded at step 45k with ray batch size 512. $(12, 2, 1, 40)$ is the default configuration.  $(12,4,1,80)^\dag$ denotes  (12, 2, 1, 40) feature configuration with the dimension of  hidden layers in the sigma MLP increased to 128. }
    \label{tab: more ablation}
    \end{table}
    %---------------------------------------

    We also provide ablation study on the structures of HybridNGP to support the superiority of the hybrid feature design. We change the latent vector dim $F$ of both spatial and temporal encodings and compare the evaluation metrics. The averaged results on DyNeRF dataset are shown in table~\ref{tab: more ablation}.    Comparing the different configurations of the sole T-Code length and the whole layout of features, we conclude that the tiny sigma MLP shown in equation~\ref{our_dnerf_sigma} with two 64-dim hidden layers is generally compatible with different configurations of feature layouts, but the quality is not optimized. The length of the sole T-Code affects the quality, but the ratio between the lengths of T-Code and spatial features are more important. 
    Increasing the length of the hybrid representations does not necessarily result in better reconstruction quality, due to the bottleneck of representation   capacity  of the 64-dim hidden layers. Better quality could be achieved by utilizing a larger MLP to solve the bottleneck.
    
    Another advantage we can conclude from table \ref{tab: more ablation} is the speed potential of our HybridNGP.  
    As is shown in table~\ref{tab: more ablation}, 
    the difference in time consumption shows that the speed of HybridNGP is not severely effected by increasing the complexities of feature from the default configuration. The results imply that the bottleneck of speed is not the complexity of latent representation: Due to the hash table  structure, the length of hybrid feature vectors only affect the computation complexity at the first layer of the downstream MLP. 
    Moreover, Due to the ray marching strategy, 3D points are sampled  according to the maintained occupancy grids, and the low batch size leads part of  computational units in the GPU to be idle.
    The bottleneck of speed  also proves the advantage and viability of  HybridNGP. On the one hand,  this model feature enables HybridNGP to cope with hybrid representations with much higher complexity  and gain quality improvement at the cost of subtle speed downgrade. On the other hand, metrics show that even if the computational resources are not exhaustively occupied, our method still yields high fidelity reconstruction. Thus, higher acceleration effects from other baselines to HybridNGP may be observed on platforms with weaker GPUs.

\subsection{DNGP-T for Monocular Dynamic View Synthesis}
\label{exp DNGP-T}
As is demonstrated in equation \ref{dngp_sigma} and \ref{dngnp_c},  we combine T-Code with deformation networks  \cite{Dnerf} and obtain DNGP-T for monocular setting to support the viability of T-Code. Following TineuVox  \cite{tineuvox} and HexPlane  \cite{HexPlane_}, we test our DNGP-T on  D-NeRF dataset  \cite{Dnerf} for evaluation. D-NeRF dataset consists of 8 synthetic dynamic scenes, with only one camera moving around. Following the work of  \cite{Dnerf} and  \cite{tineuvox}, we train and evaluate models under a resolution of 400 $\times$ 400 based on the implementation of  \cite{torch-ngp}. 
Our implementation is based on the latest commit
b6e08046 on GitHub. The spatial resolutions $N_{min}$ and $ N_{max}$ in our DNGP-T  are set as $16$ and $2048 \times$  the scale of the scene. $L, F$ for the spatial encoding are set as $12, 2$. Like \cite{HexPlane_,tineuvox}, we set different hyper-parameters of our T-Code to cater for different data distributions in different datasets. We choose T-Code with 2 levels, with $N_{min}, N_{max}$ = $30, 100$ for leveraging interpolation for DNGP-T. $T$ is set as $2^{19}$ for spatial feature, and $2^7$ for T-Code. 
 The baseline \cite{torch-ngp} and our DNGP-T are optimized by AdamW \cite{loshchilov2017decoupled}. Learning rates are  scheduled by exponential decay from 0.01 to 0.001. 
 We select $\epsilon$ in the optimizer as  $10^{-15}$. The weight decay is set as  $5\times 10^{-5}$ for the RGB MLP and our T-Code, and the default $0.01$ for other modules.  $(\beta_1,\beta_2)$  are set as $(0.9,0.99)$. The random seeds are set as 1337. 

We report average PSNR, SSIM and running time in table \ref{exp_dnerfset}. Torch-NGP  \cite{torch-ngp} is the baseline model with deformation-based spatial hash encoding shown in equation \ref{dngp_sigma} and \ref{dngnp_c}, with our T-Code  removed.  DNGP-T outperforms the others by a significant margin. Compared to the baseline Torch-NGP, DNGP-T gains an advantage in reconstruction quality at the cost of little  increase in computation. Moreover, we also provide the rendered results in figure \ref{fig:trex_comp} for qualitative  comparison.  DNGP-T outperforms the others with better details. This is because our  T-Code copes with  transient appearance change and alleviates the  disturbance from temporal information in the observations,  enabling the high resolution hash grids and deformation MLP in our DNGP-T to recover the geometry details better.

%---------------------------------------------Table------------------------------------
\begin{table}[!t]
\centering
\begin{tabular}{llll}
    \Xhline{3\arrayrulewidth}
\multicolumn{1}{l}{\textbf{Method}} & \textbf{PSNR $\uparrow$}                      & \multicolumn{1}{c}{\textbf{SSIM $\uparrow$}} & \textbf{Time $\downarrow$}        \\ \hline
D-NeRF                     & 30.50                     & 0.95                     & $\sim$hours \\
TineuVox-B                 & 32.67 & 0.97 & 28min       \\
HexPlane                   & 31.04                     & 0.97                     & \textbf{11min}       \\
Torch-NGP  \cite{torch-ngp}                  & 34.27                     & \textbf{0.98}                     & 13min       \\
DNGP-T (Ours)          & \textbf{34.95}                    & \textbf{0.98}                     & 14min      \\

\Xhline{3\arrayrulewidth}
\end{tabular}
\caption{Comparison on D-NeRF dataset. All times are recorded on a single RTX 3090 GPU.  }
\label{exp_dnerfset}
\end{table}
%------------------------------------------------------------------------------

\section{Conclusion and Limitation}

Targeting at  key difficulties for efficient dynamic view synthesis, we propose the T-Code representation for temporal features.  Our T-Code is born to be compact and precise due to the hash table structure, and the specialty in modeling temporal features enables other modules to deal with other features without confusion.  T-Code is flexible and robust enough to be combined with existing modules for dynamic view synthesis in different scenarios: HybridNGP achieves compact high fidelity multi-camera reconstruction at top speed, while DNGP-T achieves fast monocular reconstruction with state-of-the-art reconstruction quality. 

Like most approaches for dynamic scene synthesis, the training of our T-Code requires thousands of iterations across the whole dataset, posing a large burden on host memory. Further improvements will be made to accommodate for streaming settings and push dynamic view synthesis towards more  applications in real-time.

\begin{appendices}

\section{Detailed Experimental results}
\begin{figure*}[!ht]
    \centering
    \begin{subfigure}[b]{0.4\columnwidth}
        \centering
        \includegraphics[width=\textwidth]{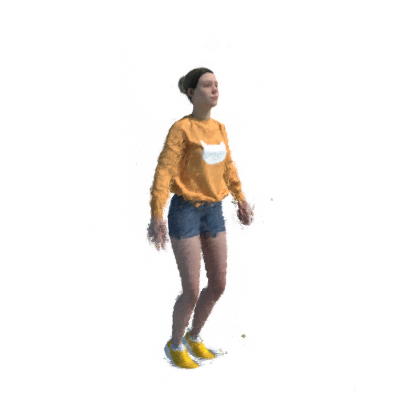}
    \caption{Torch-NGP, frame 13.}
        \label{fig:tngp13}
    \end{subfigure}
    \begin{subfigure}[b]{0.4\columnwidth}
    \centering
    \includegraphics[width=\textwidth]{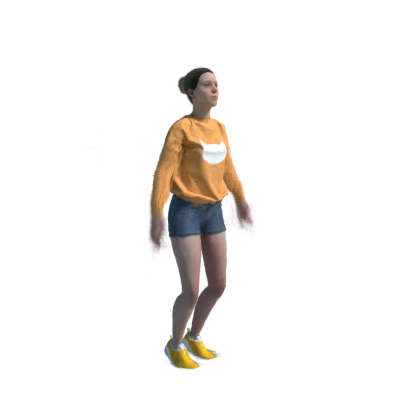}
    \caption{DNGP-T, frame 13.}
    \label{fig:dntpt13}
    \end{subfigure}
   \begin{subfigure}[b]{0.4\columnwidth}
        \centering
        \includegraphics[width=\textwidth]{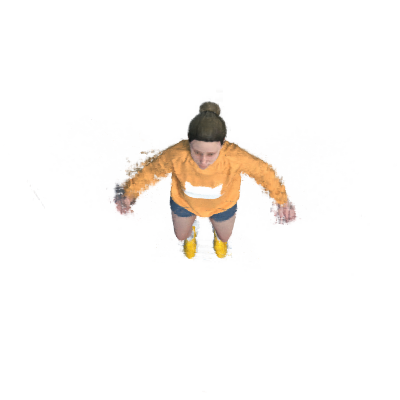}
    \caption{Torch-NGP, frame 14.}
        \label{fig:tngp14}
    \end{subfigure}
    \begin{subfigure}[b]{0.4\columnwidth}
    \centering
    \includegraphics[width=\textwidth]{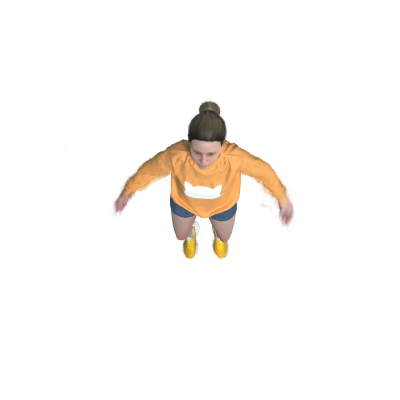}
    \caption{DNGP-T, frame 14.}
    \label{fig:dntpt14}
    \end{subfigure}

       \caption{Rendered results on the \emph{jumping jacks} scene.  }

       \label{fig:more_dnerf_jumpingjacks}
\end{figure*}

\begin{figure*}[!ht]
    \centering
    \begin{subfigure}[b]{0.4\columnwidth}
        \centering
        \includegraphics[width=\textwidth]{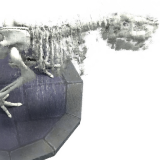}
    \caption{Torch-NGP, frame 4.}
        \label{fig:tngp4}
    \end{subfigure}
    \begin{subfigure}[b]{0.4\columnwidth}
    \centering
    \includegraphics[width=\textwidth]{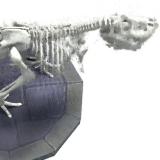}
    \caption{DNGP-T, frame 4.}
    \label{fig:dntpt4}
    \end{subfigure}
    \begin{subfigure}[b]{0.4\columnwidth}
        \centering
        \includegraphics[width=\textwidth]{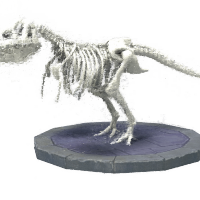}
    \caption{Torch-NGP, frame 19.}
        \label{fig:tngp19}
    \end{subfigure}
    \begin{subfigure}[b]{0.4\columnwidth}
    \centering
    \includegraphics[width=\textwidth]{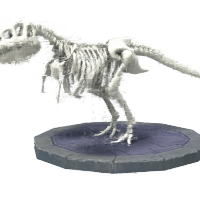}
    \caption{DNGP-T, frame 19.}
    \label{fig:dntpt19}
    \end{subfigure}
    \caption{Rendered results on the \emph{trex} scene.  }
       \label{fig:more_dnerf_trex}
\end{figure*}

\begin{figure*}[!ht]
    \centering
    \begin{subfigure}[b]{0.3\columnwidth}
        \centering
        \includegraphics[width=\textwidth]{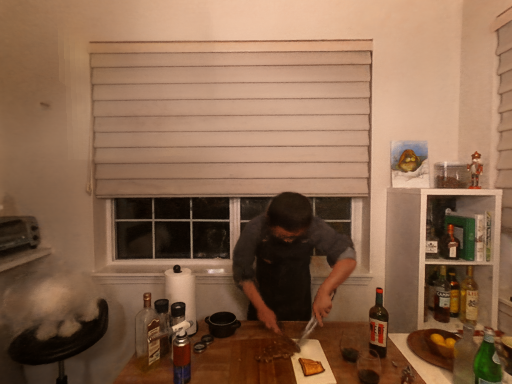}
    \end{subfigure}
    \begin{subfigure}[b]{0.3\columnwidth}
    \centering
    \includegraphics[width=\textwidth]{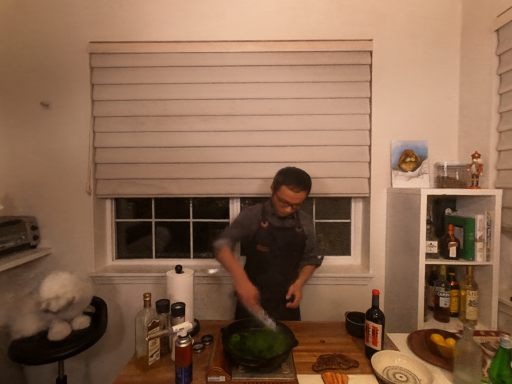}
    \end{subfigure}
   \begin{subfigure}[b]{0.3\columnwidth}
        \centering
        \includegraphics[width=\textwidth]{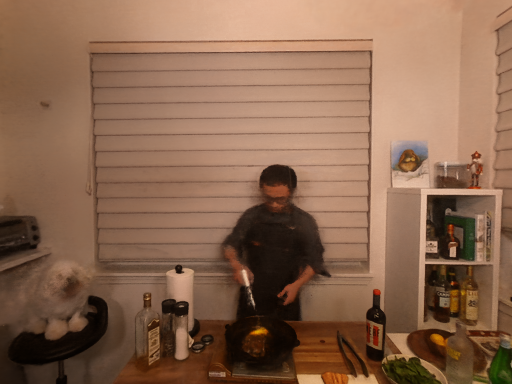}
    \end{subfigure}
    \begin{subfigure}[b]{0.3\columnwidth}
    \centering
    \includegraphics[width=\textwidth]{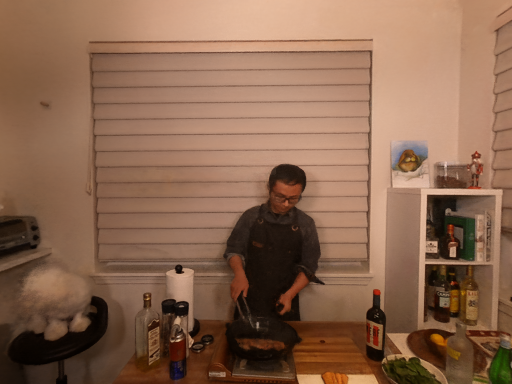}
    \end{subfigure}
    \begin{subfigure}[b]{0.3\columnwidth}
        \centering
        \includegraphics[width=\textwidth]{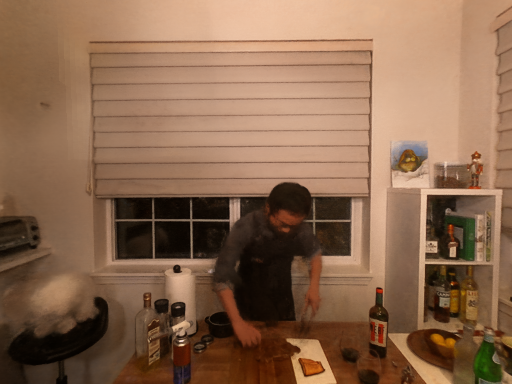}
    \end{subfigure}
    \begin{subfigure}[b]{0.3\columnwidth}
    \centering
    \includegraphics[width=\textwidth]{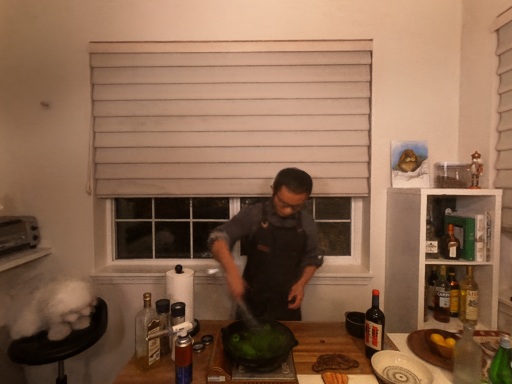}
    \end{subfigure}
    \begin{subfigure}[b]{0.3\columnwidth}
        \centering
        \includegraphics[width=\textwidth]{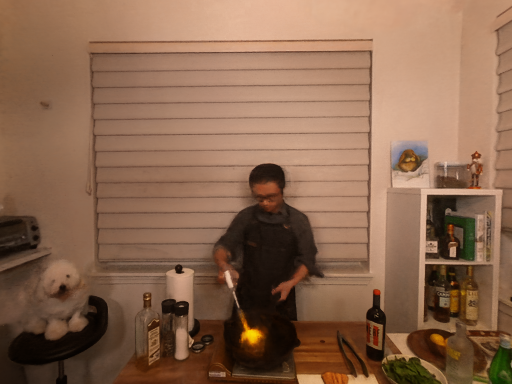}
    \end{subfigure}
    \begin{subfigure}[b]{0.3\columnwidth}
    \centering
    \includegraphics[width=\textwidth]{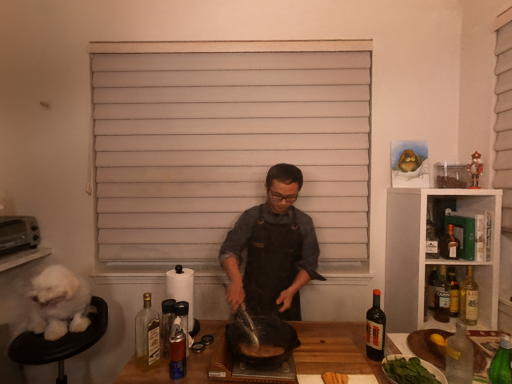}
    \end{subfigure}
    \begin{subfigure}[b]{0.3\columnwidth}
        \centering
        \includegraphics[width=\textwidth]{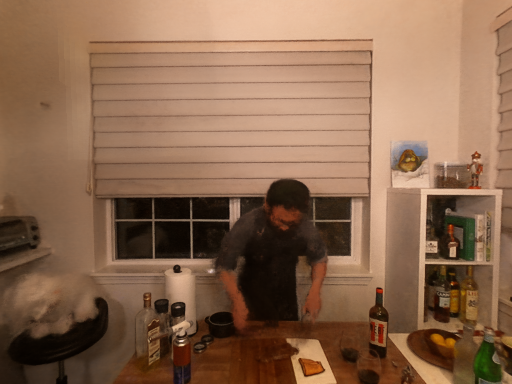}
    \end{subfigure}
    \begin{subfigure}[b]{0.3\columnwidth}
    \centering
    \includegraphics[width=\textwidth]{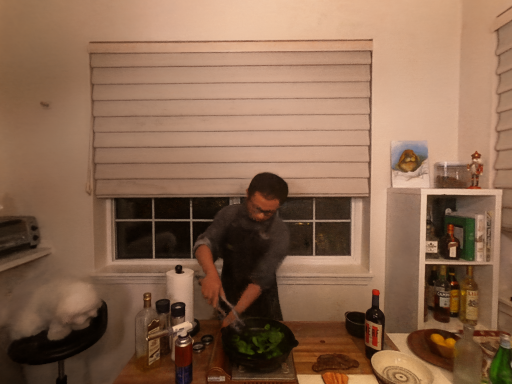}
    \end{subfigure}
    \begin{subfigure}[b]{0.3\columnwidth}
        \centering
        \includegraphics[width=\textwidth]{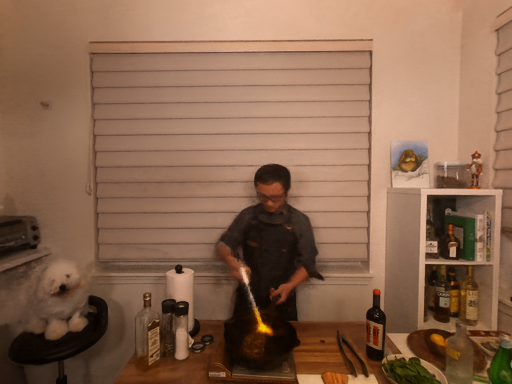}
    \end{subfigure}
    \begin{subfigure}[b]{0.3\columnwidth}
    \centering
    \includegraphics[width=\textwidth]{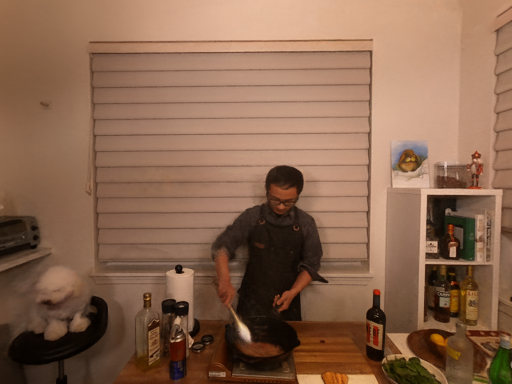}
    \end{subfigure}

   \caption{Rendered results of HybridNGP with default settings on \emph{DyNeRF dataset}. We provide results at step 90k.  }
   \label{fig: dynerf details}
\end{figure*}

 We report per-scene results of HybridNGP in table~\ref{tab:perscene_dynerf} .   We also report more rendered images in DyNeRF dataset in figure  \ref{fig: dynerf details}. 
 For DNGP-T, 
 we report per-scene metrics in table~\ref{tab:per-scene Dnerf}. We also report more comparisons of baseline \cite{torch-ngp} and DNGP-T on rendered images  in D-NeRF dataset  in figure~\ref{fig:more_dnerf_jumpingjacks} and \ref{fig:more_dnerf_trex}. As demonstrated, the proposed DNGP-T  generates images with much less artifacts.

%------------------table-detailed-Dynerf
\begin{table*}[!ht]
    \centering
    \begin{tabular}{l|cccccc}
        \Xhline{3\arrayrulewidth}
    \textbf{Steps}         & \multicolumn{3}{c}{\textbf{Flame Salmon}}            & \multicolumn{3}{c}{\textbf{Cook Spinach}}          \\
                 & \textbf{PSNR $\uparrow$}&\textbf{DSSIM $\downarrow$}& \multicolumn{1}{l|}{\textbf{LPIPS $\downarrow$}} & \textbf{PSNR $\uparrow$}       & \textbf{DSSIM $\downarrow$}      & \textbf{LPIPS $\downarrow$}      \\ \hline
    45k & 28.58 & 0.0172 & 0.0753                     & 32.02 & 0.0133 & 0.0667                        \\
    90k & 28.67 & 0.0167 & 0.0699                     & 32.18 & 0.0128 & 0.0628                         \\
    \Xhline{3\arrayrulewidth}
    
    \textbf{Steps}         & \multicolumn{3}{c}{\textbf{Flame Steak}}            & \multicolumn{3}{c}{\textbf{Sear Steak}}                     \\
                  &\textbf{PSNR $\uparrow$}&\textbf{DSSIM $\downarrow$}& \multicolumn{1}{l|}{\textbf{LPIPS $\downarrow$} } & \textbf{PSNR $\uparrow$}       & \textbf{DSSIM $\downarrow$}     & \textbf{LPIPS $\downarrow$}      \\ \hline
    45k & 31.74 & 0.0119 & 0.0629                     & 33.01 & 0.0089 & 0.0498                        \\
    90k & 32.24 & 0.0109 & 0.0573                     & 33.27 & 0.0086 & 0.0460                       \\    \Xhline{3\arrayrulewidth}
    
    \textbf{Steps}         & \multicolumn{3}{c}{\textbf{Cut Roasted Beef}}                    & \multicolumn{3}{c}{\textbf{Average}}          \\
                  &\textbf{PSNR $\uparrow$}&\textbf{DSSIM $\downarrow$}& \multicolumn{1}{l|}{\textbf{LPIPS $\downarrow$} } & \textbf{PSNR $\uparrow$}       & \textbf{DSSIM $\downarrow$}     & \textbf{LPIPS $\downarrow$}      \\ \hline
    45k & 31.69      & 0.0141     & 0.0728                                        & 31.41      & 0.0131     & 0.0655     \\
    90k & 31.86      & 0.0142     & 0.0690                                        & 31.64      & 0.0126     & 0.0609     \\    \Xhline{3\arrayrulewidth}
    \end{tabular}
    \caption{Detailed results of HybridNGP on each scene of DyNeRF dataset.  \emph{Steps} denotes the number of iterations performed by Adam\cite{adamoptim} optimizer. }
    \label{tab:perscene_dynerf}
    \end{table*}
    %-----------------------------------------

%---------------------figure: dnerf-details

%--------------------------------------------

%------------------table-detailed-dnerf
\begin{table*}[htbp]
    \centering
    \begin{tabular}{l|llllll} 
        \Xhline{3\arrayrulewidth}
    \textbf{Model} & \multicolumn{2}{c}{\textbf{Hell Warrior}}                                                                                              & \multicolumn{2}{c}{\textbf{Mutant}}                                                                                                    & \multicolumn{2}{c}{\textbf{Hook}}                                                                                                      \\
                                    & \multicolumn{1}{c}{\textbf{PSNR $\uparrow$}} & \multicolumn{1}{c}{\textbf{SSIM $\uparrow$}} & \multicolumn{1}{c}{\textbf{PSNR $\uparrow$}} & \multicolumn{1}{c}{\textbf{SSIM $\uparrow$}} & \multicolumn{1}{c}{\textbf{PSNR $\uparrow$}} & \multicolumn{1}{c}{\textbf{SSIM $\uparrow$}} \\ \hline
    
    D-NeRF                          & 25.02                                                         & 0.95                                                          & 31.29                                                         & 0.97                                                          & 29.25                                                         & 0.96                                                          \\
    
    TiNeuVox-S                      & 27                                                            & 0.95                                                          & 31.09                                                         & 0.96                                                          & 29.3                                                          & 0.95                                                          \\
    
    TiNeuVox-B                      & 28.17                                                         & 0.97                                                          & 33.61                                                         & 0.98                                                          & 31.45                                                         & 0.97                                                          \\
    
    HexPlane                        & 24.24                                                         & 0.94                                                          & 33.79                                                         & 0.98                                                          & 28.71                                                         & 0.96                                                          \\
    Torch-NGP                       & 28.57                                 & 0.97                                                          & 39.35                                                         & 0.99                                                          & 33.74                                                         & 0.98                                                          \\
    DNGP-T                          & 28.84                                 & 0.97                                                          & 39.30                                                         & 0.99                                                          & 34.11                                                         & 0.98                                                          \\    \Xhline{3\arrayrulewidth}
    \textbf{Model} & \multicolumn{2}{c}{\textbf{Bouncing Balls}}                                                                                            & \multicolumn{2}{c}{\textbf{Lego}}                                                                                                      & \multicolumn{2}{c}{\textbf{Trex}}                                                                                                      \\
                                    & \multicolumn{1}{c}{\textbf{PSNR $\uparrow$}} & \multicolumn{1}{c}{\textbf{SSIM $\uparrow$}} & \multicolumn{1}{c}{\textbf{PSNR $\uparrow$}} & \multicolumn{1}{c}{\textbf{SSIM $\uparrow$}} & \multicolumn{1}{c}{\textbf{PSNR $\uparrow$}} & \multicolumn{1}{c}{\textbf{SSIM $\uparrow$}} \\ \hline
    
    D-NeRF                          & 38.93                                                         & 0.98                                                          & 21.64                                                         & 0.83                                                          & 31.75                                                         & 0.97                                                          \\
    
    TiNeuVox-S                      & 39.05                                                         & 0.99                                                          & 24.35                                                         & 0.88                                                          & 29.95                                                         & 0.96                                                          \\
    
    TiNeuVox-B                      & 40.73                                                         & 0.99                                                          & 25.02                                                         & 0.92                                                          & 32.7                                                          & 0.98                                                          \\
    
    HexPlane                        & 39.69                                                         & 0.99                                                          & 25.22                                                         & 0.94                                                          & 30.67                                                         & 0.98                                                          \\
    Torch-NGP                       & 41.64                                 & 0.99                                                          & 25.39                                                         & 0.94                                                          & 34.66                                                         & 0.98                                                          \\
    DNGP-T                          & 41.39                                 & 0.99                                                          & 25.27                                                         & 0.94                                                          & 36.93                                                         & 0.99                                                          \\    \Xhline{3\arrayrulewidth}
    \textbf{Model} & \multicolumn{2}{c}{\textbf{Stand up}}                                                                                                  & \multicolumn{2}{c}{\textbf{Jumping Jacks}}                                                                                             & \multicolumn{2}{c}{\textbf{Average}}                                                                                                   \\
                                    & \multicolumn{1}{c}{\textbf{PSNR $\uparrow$}} & \multicolumn{1}{c}{\textbf{SSIM $\uparrow$}} & \multicolumn{1}{c}{\textbf{PSNR $\uparrow$}} & \multicolumn{1}{c}{\textbf{SSIM $\uparrow$}} & \multicolumn{1}{c}{\textbf{PSNR $\uparrow$}} & \multicolumn{1}{c}{\textbf{SSIM $\uparrow$}} \\\hline
    D-NeRF                          & 32.79                                                         & 0.98                                                          & 32.8                                                          & 0.98                                                          & 30.5                                                          & 0.95                                                          \\
    
    TiNeuVox-S                      & 32.89                                                         & 0.98                                                          & 32.33                                                         & 0.97                                                          & 30.75                                                         & 0.96                                                          \\
    
    TiNeuVox-B                      & 35.43                                                         & 0.99                                                          & 34.23                                                         & 0.98                                                          & 32.64                                                         & 0.97                                                          \\
    
    HexPlane                        & 34.36                                                         & 0.98                                                          & 31.65                                                         & 0.97                                                          & 31.04                                                         & 0.97                                                          \\
    Torch-NGP                       & 38.91                                 & 0.99                                                          & 31.86                                                         & 0.97                                                          & 34.27                                                         & 0.98                                                          \\
    DNGP-T                          & 38.82                                 & 0.99                                                          & 34.96                                                         & 0.99                                                          & 34.95                                                         & 0.98                                               \\     \Xhline{3\arrayrulewidth}
    \end{tabular}
    \caption{Comparison of per-scene results of  our DNGP-T and previous methods on D-NeRF dataset. We report metrics of baseline  \emph{Torch-NGP} \cite{torch-ngp} and our DNGP-T.    The metrics of the other methods are copied from \cite{HexPlane_}. }
    \label{tab:per-scene Dnerf}
    \end{table*}
%---------------------------------------

\end{appendices}

%%===========================================================================================%%
%% If you are submitting to one of the Nature Portfolio journals, using the eJP submission   %%
%% system, please include the references within the manuscript file itself. You may do this  %%
%% by copying the reference list from your .bbl file, paste it into the main manuscript .tex %%
%% file, and delete the associated \verb+\bibliography+ commands.                            %%
%%===========================================================================================%%

\newpage

\bibliography{sn-article.bbl}% common bib file

%% BioMed_Central_Bib_Style_v1.01

\begin{thebibliography}{32}
% BibTex style file: bmc-mathphys.bst (version 2.1), 2014-07-24
\ifx \bisbn   \undefined \def \bisbn  #1{ISBN #1}\fi
\ifx \binits  \undefined \def \binits#1{#1}\fi
\ifx \bauthor  \undefined \def \bauthor#1{#1}\fi
\ifx \batitle  \undefined \def \batitle#1{#1}\fi
\ifx \bjtitle  \undefined \def \bjtitle#1{#1}\fi
\ifx \bvolume  \undefined \def \bvolume#1{\textbf{#1}}\fi
\ifx \byear  \undefined \def \byear#1{#1}\fi
\ifx \bissue  \undefined \def \bissue#1{#1}\fi
\ifx \bfpage  \undefined \def \bfpage#1{#1}\fi
\ifx \blpage  \undefined \def \blpage #1{#1}\fi
\ifx \burl  \undefined \def \burl#1{\textsf{#1}}\fi
\ifx \doiurl  \undefined \def \doiurl#1{\url{https://doi.org/#1}}\fi
\ifx \betal  \undefined \def \betal{\textit{et al.}}\fi
\ifx \binstitute  \undefined \def \binstitute#1{#1}\fi
\ifx \binstitutionaled  \undefined \def \binstitutionaled#1{#1}\fi
\ifx \bctitle  \undefined \def \bctitle#1{#1}\fi
\ifx \beditor  \undefined \def \beditor#1{#1}\fi
\ifx \bpublisher  \undefined \def \bpublisher#1{#1}\fi
\ifx \bbtitle  \undefined \def \bbtitle#1{#1}\fi
\ifx \bedition  \undefined \def \bedition#1{#1}\fi
\ifx \bseriesno  \undefined \def \bseriesno#1{#1}\fi
\ifx \blocation  \undefined \def \blocation#1{#1}\fi
\ifx \bsertitle  \undefined \def \bsertitle#1{#1}\fi
\ifx \bsnm \undefined \def \bsnm#1{#1}\fi
\ifx \bsuffix \undefined \def \bsuffix#1{#1}\fi
\ifx \bparticle \undefined \def \bparticle#1{#1}\fi
\ifx \barticle \undefined \def \barticle#1{#1}\fi
\bibcommenthead
\ifx \bconfdate \undefined \def \bconfdate #1{#1}\fi
\ifx \botherref \undefined \def \botherref #1{#1}\fi
\ifx \url \undefined \def \url#1{\textsf{#1}}\fi
\ifx \bchapter \undefined \def \bchapter#1{#1}\fi
\ifx \bbook \undefined \def \bbook#1{#1}\fi
\ifx \bcomment \undefined \def \bcomment#1{#1}\fi
\ifx \oauthor \undefined \def \oauthor#1{#1}\fi
\ifx \citeauthoryear \undefined \def \citeauthoryear#1{#1}\fi
\ifx \endbibitem  \undefined \def \endbibitem {}\fi
\ifx \bconflocation  \undefined \def \bconflocation#1{#1}\fi
\ifx \arxivurl  \undefined \def \arxivurl#1{\textsf{#1}}\fi
\csname PreBibitemsHook\endcsname

%%% 1
\bibitem[\protect\citeauthoryear{Mildenhall et~al.}{2020}]{nerfeccv}
\begin{bchapter}
\bauthor{\bsnm{Mildenhall}, \binits{B.}},
\bauthor{\bsnm{Srinivasan}, \binits{P.P.}},
\bauthor{\bsnm{Tancik}, \binits{M.}},
\bauthor{\bsnm{Barron}, \binits{J.T.}},
\bauthor{\bsnm{Ramamoorthi}, \binits{R.}},
\bauthor{\bsnm{Ng}, \binits{R.}}:
\bctitle{Nerf: Representing scenes as neural radiance fields for view synthesis}.
In: \beditor{\bsnm{Vedaldi}, \binits{A.}},
\beditor{\bsnm{Bischof}, \binits{H.}},
\beditor{\bsnm{Brox}, \binits{T.}},
\beditor{\bsnm{Frahm}, \binits{J.-M.}} (eds.)
\bbtitle{Computer Vision -- ECCV 2020},
pp. \bfpage{405}--\blpage{421}.
\bpublisher{Springer},
\blocation{Cham}
(\byear{2020})
\end{bchapter}
\endbibitem

%%% 2
\bibitem[\protect\citeauthoryear{Lombardi et~al.}{2019}]{neuralvolume}
\begin{botherref}
\oauthor{\bsnm{Lombardi}, \binits{S.}},
\oauthor{\bsnm{Simon}, \binits{T.}},
\oauthor{\bsnm{Saragih}, \binits{J.}},
\oauthor{\bsnm{Schwartz}, \binits{G.}},
\oauthor{\bsnm{Lehrmann}, \binits{A.}},
\oauthor{\bsnm{Sheikh}, \binits{Y.}}:
Neural volumes: Learning dynamic renderable volumes from images.
arXiv preprint arXiv:1906.07751
(2019)
\end{botherref}
\endbibitem

%%% 3
\bibitem[\protect\citeauthoryear{Mildenhall et~al.}{2019}]{llff}
\begin{barticle}
\bauthor{\bsnm{Mildenhall}, \binits{B.}},
\bauthor{\bsnm{Srinivasan}, \binits{P.P.}},
\bauthor{\bsnm{Ortiz-Cayon}, \binits{R.}},
\bauthor{\bsnm{Kalantari}, \binits{N.K.}},
\bauthor{\bsnm{Ramamoorthi}, \binits{R.}},
\bauthor{\bsnm{Ng}, \binits{R.}},
\bauthor{\bsnm{Kar}, \binits{A.}}:
\batitle{Local light field fusion: Practical view synthesis with prescriptive sampling guidelines}.
\bjtitle{ACM Transactions on Graphics (TOG)}
\bvolume{38}(\bissue{4}),
\bfpage{1}--\blpage{14}
(\byear{2019})
\end{barticle}
\endbibitem

%%% 4
\bibitem[\protect\citeauthoryear{Barron et~al.}{2021}]{Mipnerf}
\begin{bchapter}
\bauthor{\bsnm{Barron}, \binits{J.T.}},
\bauthor{\bsnm{Mildenhall}, \binits{B.}},
\bauthor{\bsnm{Tancik}, \binits{M.}},
\bauthor{\bsnm{Hedman}, \binits{P.}},
\bauthor{\bsnm{Martin-Brualla}, \binits{R.}},
\bauthor{\bsnm{Srinivasan}, \binits{P.P.}}:
\bctitle{Mip-nerf: A multiscale representation for anti-aliasing neural radiance fields}.
In: \bbtitle{Proceedings of the IEEE/CVF International Conference on Computer Vision (ICCV)},
pp. \bfpage{5855}--\blpage{5864}
(\byear{2021})
\end{bchapter}
\endbibitem

%%% 5
\bibitem[\protect\citeauthoryear{Sabour et~al.}{2023}]{robustnerf}
\begin{bchapter}
\bauthor{\bsnm{Sabour}, \binits{S.}},
\bauthor{\bsnm{Vora}, \binits{S.}},
\bauthor{\bsnm{Duckworth}, \binits{D.}},
\bauthor{\bsnm{Krasin}, \binits{I.}},
\bauthor{\bsnm{Fleet}, \binits{D.J.}},
\bauthor{\bsnm{Tagliasacchi}, \binits{A.}}:
\bctitle{Robustnerf: Ignoring distractors with robust losses}.
In: \bbtitle{Proceedings of the IEEE/CVF Conference on Computer Vision and Pattern Recognition (CVPR)},
pp. \bfpage{20626}--\blpage{20636}
(\byear{2023})
\end{bchapter}
\endbibitem

%%% 6
\bibitem[\protect\citeauthoryear{Niemeyer et~al.}{2022}]{regnerf}
\begin{bchapter}
\bauthor{\bsnm{Niemeyer}, \binits{M.}},
\bauthor{\bsnm{Barron}, \binits{J.T.}},
\bauthor{\bsnm{Mildenhall}, \binits{B.}},
\bauthor{\bsnm{Sajjadi}, \binits{M.S.}},
\bauthor{\bsnm{Geiger}, \binits{A.}},
\bauthor{\bsnm{Radwan}, \binits{N.}}:
\bctitle{Regnerf: Regularizing neural radiance fields for view synthesis from sparse inputs}.
In: \bbtitle{Proceedings of the IEEE/CVF Conference on Computer Vision and Pattern Recognition},
pp. \bfpage{5480}--\blpage{5490}
(\byear{2022})
\end{bchapter}
\endbibitem

%%% 7
\bibitem[\protect\citeauthoryear{Deng et~al.}{2022}]{dsnerf}
\begin{bchapter}
\bauthor{\bsnm{Deng}, \binits{K.}},
\bauthor{\bsnm{Liu}, \binits{A.}},
\bauthor{\bsnm{Zhu}, \binits{J.-Y.}},
\bauthor{\bsnm{Ramanan}, \binits{D.}}:
\bctitle{Depth-supervised nerf: Fewer views and faster training for free}.
In: \bbtitle{Proceedings of the IEEE/CVF Conference on Computer Vision and Pattern Recognition},
pp. \bfpage{12882}--\blpage{12891}
(\byear{2022})
\end{bchapter}
\endbibitem

%%% 8
\bibitem[\protect\citeauthoryear{Fridovich-Keil et~al.}{2022}]{plenoxels}
\begin{bchapter}
\bauthor{\bsnm{Fridovich-Keil}, \binits{S.}},
\bauthor{\bsnm{Yu}, \binits{A.}},
\bauthor{\bsnm{Tancik}, \binits{M.}},
\bauthor{\bsnm{Chen}, \binits{Q.}},
\bauthor{\bsnm{Recht}, \binits{B.}},
\bauthor{\bsnm{Kanazawa}, \binits{A.}}:
\bctitle{Plenoxels: Radiance fields without neural networks}.
In: \bbtitle{CVPR}
(\byear{2022})
\end{bchapter}
\endbibitem

%%% 9
\bibitem[\protect\citeauthoryear{Reiser et~al.}{2021}]{kilonerf}
\begin{bchapter}
\bauthor{\bsnm{Reiser}, \binits{C.}},
\bauthor{\bsnm{Peng}, \binits{S.}},
\bauthor{\bsnm{Liao}, \binits{Y.}},
\bauthor{\bsnm{Geiger}, \binits{A.}}:
\bctitle{Kilonerf: Speeding up neural radiance fields with thousands of tiny mlps}.
In: \bbtitle{International Conference on Computer Vision (ICCV)}
(\byear{2021})
\end{bchapter}
\endbibitem

%%% 10
\bibitem[\protect\citeauthoryear{Yang et~al.}{2022}]{recursivenerf}
\begin{botherref}
\oauthor{\bsnm{Yang}, \binits{G.-W.}},
\oauthor{\bsnm{Zhou}, \binits{W.-Y.}},
\oauthor{\bsnm{Peng}, \binits{H.-Y.}},
\oauthor{\bsnm{Liang}, \binits{D.}},
\oauthor{\bsnm{Mu}, \binits{T.-J.}},
\oauthor{\bsnm{Hu}, \binits{S.-M.}}:
Recursive-nerf: An efficient and dynamically growing nerf.
IEEE Transactions on Visualization and Computer Graphics
(2022)
\end{botherref}
\endbibitem

%%% 11
\bibitem[\protect\citeauthoryear{Schonberger and Frahm}{2016}]{sfmotion}
\begin{bchapter}
\bauthor{\bsnm{Schonberger}, \binits{J.L.}},
\bauthor{\bsnm{Frahm}, \binits{J.-M.}}:
\bctitle{Structure-from-motion revisited}.
In: \bbtitle{Proceedings of the IEEE Conference on Computer Vision and Pattern Recognition},
pp. \bfpage{4104}--\blpage{4113}
(\byear{2016})
\end{bchapter}
\endbibitem

%%% 12
\bibitem[\protect\citeauthoryear{M{\"u}ller et~al.}{2022}]{ingp}
\begin{barticle}
\bauthor{\bsnm{M{\"u}ller}, \binits{T.}},
\bauthor{\bsnm{Evans}, \binits{A.}},
\bauthor{\bsnm{Schied}, \binits{C.}},
\bauthor{\bsnm{Keller}, \binits{A.}}:
\batitle{Instant neural graphics primitives with a multiresolution hash encoding}.
\bjtitle{ACM Transactions on Graphics (ToG)}
\bvolume{41}(\bissue{4}),
\bfpage{1}--\blpage{15}
(\byear{2022})
\end{barticle}
\endbibitem

%%% 13
\bibitem[\protect\citeauthoryear{Yariv et~al.}{2023}]{yariv2023bakedsdf}
\begin{botherref}
\oauthor{\bsnm{Yariv}, \binits{L.}},
\oauthor{\bsnm{Hedman}, \binits{P.}},
\oauthor{\bsnm{Reiser}, \binits{C.}},
\oauthor{\bsnm{Verbin}, \binits{D.}},
\oauthor{\bsnm{Srinivasan}, \binits{P.P.}},
\oauthor{\bsnm{Szeliski}, \binits{R.}},
\oauthor{\bsnm{Barron}, \binits{J.T.}},
\oauthor{\bsnm{Mildenhall}, \binits{B.}}:
Bakedsdf: Meshing neural sdfs for real-time view synthesis.
arXiv preprint arXiv:2302.14859
(2023)
\end{botherref}
\endbibitem

%%% 14
\bibitem[\protect\citeauthoryear{Yuan et~al.}{2022}]{editnerf}
\begin{bchapter}
\bauthor{\bsnm{Yuan}, \binits{Y.-J.}},
\bauthor{\bsnm{Sun}, \binits{Y.-T.}},
\bauthor{\bsnm{Lai}, \binits{Y.-K.}},
\bauthor{\bsnm{Ma}, \binits{Y.}},
\bauthor{\bsnm{Jia}, \binits{R.}},
\bauthor{\bsnm{Gao}, \binits{L.}}:
\bctitle{Nerf-editing: geometry editing of neural radiance fields}.
In: \bbtitle{Proceedings of the IEEE/CVF Conference on Computer Vision and Pattern Recognition},
pp. \bfpage{18353}--\blpage{18364}
(\byear{2022})
\end{bchapter}
\endbibitem

%%% 15
\bibitem[\protect\citeauthoryear{Barron et~al.}{2022}]{mipnerf360}
\begin{bchapter}
\bauthor{\bsnm{Barron}, \binits{J.T.}},
\bauthor{\bsnm{Mildenhall}, \binits{B.}},
\bauthor{\bsnm{Verbin}, \binits{D.}},
\bauthor{\bsnm{Srinivasan}, \binits{P.P.}},
\bauthor{\bsnm{Hedman}, \binits{P.}}:
\bctitle{Mip-nerf 360: Unbounded anti-aliased neural radiance fields}.
In: \bbtitle{Proceedings of the IEEE/CVF Conference on Computer Vision and Pattern Recognition (CVPR)},
pp. \bfpage{5470}--\blpage{5479}
(\byear{2022})
\end{bchapter}
\endbibitem

%%% 16
\bibitem[\protect\citeauthoryear{Wang et~al.}{2023}]{F2nerf}
\begin{bchapter}
\bauthor{\bsnm{Wang}, \binits{P.}},
\bauthor{\bsnm{Liu}, \binits{Y.}},
\bauthor{\bsnm{Chen}, \binits{Z.}},
\bauthor{\bsnm{Liu}, \binits{L.}},
\bauthor{\bsnm{Liu}, \binits{Z.}},
\bauthor{\bsnm{Komura}, \binits{T.}},
\bauthor{\bsnm{Theobalt}, \binits{C.}},
\bauthor{\bsnm{Wang}, \binits{W.}}:
\bctitle{F2-nerf: Fast neural radiance field training with free camera trajectories}.
In: \bbtitle{Proceedings of the IEEE/CVF Conference on Computer Vision and Pattern Recognition (CVPR)},
pp. \bfpage{4150}--\blpage{4159}
(\byear{2023})
\end{bchapter}
\endbibitem

%%% 17
\bibitem[\protect\citeauthoryear{Turki et~al.}{2023}]{suds}
\begin{bchapter}
\bauthor{\bsnm{Turki}, \binits{H.}},
\bauthor{\bsnm{Zhang}, \binits{J.Y.}},
\bauthor{\bsnm{Ferroni}, \binits{F.}},
\bauthor{\bsnm{Ramanan}, \binits{D.}}:
\bctitle{Suds: Scalable urban dynamic scenes}.
In: \bbtitle{Proceedings of the IEEE/CVF Conference on Computer Vision and Pattern Recognition},
pp. \bfpage{12375}--\blpage{12385}
(\byear{2023})
\end{bchapter}
\endbibitem

%%% 18
\bibitem[\protect\citeauthoryear{Chen et~al.}{2022}]{tensorf}
\begin{bchapter}
\bauthor{\bsnm{Chen}, \binits{A.}},
\bauthor{\bsnm{Xu}, \binits{Z.}},
\bauthor{\bsnm{Geiger}, \binits{A.}},
\bauthor{\bsnm{Yu}, \binits{J.}},
\bauthor{\bsnm{Su}, \binits{H.}}:
\bctitle{Tensorf: Tensorial radiance fields}.
In: \bbtitle{European Conference on Computer Vision},
pp. \bfpage{333}--\blpage{350}
(\byear{2022}).
\bcomment{Springer}
\end{bchapter}
\endbibitem

%%% 19
\bibitem[\protect\citeauthoryear{Li et~al.}{2021}]{nsff}
\begin{bchapter}
\bauthor{\bsnm{Li}, \binits{Z.}},
\bauthor{\bsnm{Niklaus}, \binits{S.}},
\bauthor{\bsnm{Snavely}, \binits{N.}},
\bauthor{\bsnm{Wang}, \binits{O.}}:
\bctitle{Neural scene flow fields for space-time view synthesis of dynamic scenes}.
In: \bbtitle{Proceedings of the IEEE/CVF Conference on Computer Vision and Pattern Recognition},
pp. \bfpage{6498}--\blpage{6508}
(\byear{2021})
\end{bchapter}
\endbibitem

%%% 20
\bibitem[\protect\citeauthoryear{Pumarola et~al.}{2021}]{Dnerf}
\begin{bchapter}
\bauthor{\bsnm{Pumarola}, \binits{A.}},
\bauthor{\bsnm{Corona}, \binits{E.}},
\bauthor{\bsnm{Pons-Moll}, \binits{G.}},
\bauthor{\bsnm{Moreno-Noguer}, \binits{F.}}:
\bctitle{D-nerf: Neural radiance fields for dynamic scenes}.
In: \bbtitle{Proceedings of the IEEE/CVF Conference on Computer Vision and Pattern Recognition (CVPR)},
pp. \bfpage{10318}--\blpage{10327}
(\byear{2021})
\end{bchapter}
\endbibitem

%%% 21
\bibitem[\protect\citeauthoryear{Li et~al.}{2022}]{Dynerf}
\begin{bchapter}
\bauthor{\bsnm{Li}, \binits{T.}},
\bauthor{\bsnm{Slavcheva}, \binits{M.}},
\bauthor{\bsnm{Zollh\"ofer}, \binits{M.}},
\bauthor{\bsnm{Green}, \binits{S.}},
\bauthor{\bsnm{Lassner}, \binits{C.}},
\bauthor{\bsnm{Kim}, \binits{C.}},
\bauthor{\bsnm{Schmidt}, \binits{T.}},
\bauthor{\bsnm{Lovegrove}, \binits{S.}},
\bauthor{\bsnm{Goesele}, \binits{M.}},
\bauthor{\bsnm{Newcombe}, \binits{R.}},
\bauthor{\bsnm{Lv}, \binits{Z.}}:
\bctitle{Neural 3d video synthesis from multi-view video}.
In: \bbtitle{Proceedings of the IEEE/CVF Conference on Computer Vision and Pattern Recognition (CVPR)},
pp. \bfpage{5521}--\blpage{5531}
(\byear{2022})
\end{bchapter}
\endbibitem

%%% 22
\bibitem[\protect\citeauthoryear{Cao and Johnson}{2023}]{HexPlane_}
\begin{botherref}
\oauthor{\bsnm{Cao}, \binits{A.}},
\oauthor{\bsnm{Johnson}, \binits{J.}}:
Hexplane: A fast representation for dynamic scenes.
CVPR
(2023)
\end{botherref}
\endbibitem

%%% 23
\bibitem[\protect\citeauthoryear{Fridovich-Keil et~al.}{2023}]{kplanes}
\begin{bchapter}
\bauthor{\bsnm{Fridovich-Keil}, \binits{S.}},
\bauthor{\bsnm{Meanti}, \binits{G.}},
\bauthor{\bsnm{Warburg}, \binits{F.R.}},
\bauthor{\bsnm{Recht}, \binits{B.}},
\bauthor{\bsnm{Kanazawa}, \binits{A.}}:
\bctitle{K-planes: Explicit radiance fields in space, time, and appearance}.
In: \bbtitle{Proceedings of the IEEE/CVF Conference on Computer Vision and Pattern Recognition},
pp. \bfpage{12479}--\blpage{12488}
(\byear{2023})
\end{bchapter}
\endbibitem

%%% 24
\bibitem[\protect\citeauthoryear{Fang et~al.}{2022}]{tineuvox}
\begin{bchapter}
\bauthor{\bsnm{Fang}, \binits{J.}},
\bauthor{\bsnm{Yi}, \binits{T.}},
\bauthor{\bsnm{Wang}, \binits{X.}},
\bauthor{\bsnm{Xie}, \binits{L.}},
\bauthor{\bsnm{Zhang}, \binits{X.}},
\bauthor{\bsnm{Liu}, \binits{W.}},
\bauthor{\bsnm{Nie{\ss}ner}, \binits{M.}},
\bauthor{\bsnm{Tian}, \binits{Q.}}:
\bctitle{Fast dynamic radiance fields with time-aware neural voxels}.
In: \bbtitle{SIGGRAPH Asia 2022 Conference Papers},
pp. \bfpage{1}--\blpage{9}
(\byear{2022})
\end{bchapter}
\endbibitem

%%% 25
\bibitem[\protect\citeauthoryear{Park et~al.}{2023}]{park2023temporal}
\begin{bchapter}
\bauthor{\bsnm{Park}, \binits{S.}},
\bauthor{\bsnm{Son}, \binits{M.}},
\bauthor{\bsnm{Jang}, \binits{S.}},
\bauthor{\bsnm{Ahn}, \binits{Y.C.}},
\bauthor{\bsnm{Kim}, \binits{J.-Y.}},
\bauthor{\bsnm{Kang}, \binits{N.}}:
\bctitle{Temporal interpolation is all you need for dynamic neural radiance fields}.
In: \bbtitle{Proceedings of the IEEE/CVF Conference on Computer Vision and Pattern Recognition},
pp. \bfpage{4212}--\blpage{4221}
(\byear{2023})
\end{bchapter}
\endbibitem

%%% 26
\bibitem[\protect\citeauthoryear{Li et~al.}{2022}]{streamrf}
\begin{barticle}
\bauthor{\bsnm{Li}, \binits{L.}},
\bauthor{\bsnm{Shen}, \binits{Z.}},
\bauthor{\bsnm{Wang}, \binits{Z.}},
\bauthor{\bsnm{Shen}, \binits{L.}},
\bauthor{\bsnm{Tan}, \binits{P.}}:
\batitle{Streaming radiance fields for 3d video synthesis}.
\bjtitle{Advances in Neural Information Processing Systems}
\bvolume{35},
\bfpage{13485}--\blpage{13498}
(\byear{2022})
\end{barticle}
\endbibitem

%%% 27
\bibitem[\protect\citeauthoryear{Wang et~al.}{2023}]{rerf}
\begin{bchapter}
\bauthor{\bsnm{Wang}, \binits{L.}},
\bauthor{\bsnm{Hu}, \binits{Q.}},
\bauthor{\bsnm{He}, \binits{Q.}},
\bauthor{\bsnm{Wang}, \binits{Z.}},
\bauthor{\bsnm{Yu}, \binits{J.}},
\bauthor{\bsnm{Tuytelaars}, \binits{T.}},
\bauthor{\bsnm{Xu}, \binits{L.}},
\bauthor{\bsnm{Wu}, \binits{M.}}:
\bctitle{Neural residual radiance fields for streamably free-viewpoint videos}.
In: \bbtitle{Proceedings of the IEEE/CVF Conference on Computer Vision and Pattern Recognition},
pp. \bfpage{76}--\blpage{87}
(\byear{2023})
\end{bchapter}
\endbibitem

%%% 28
\bibitem[\protect\citeauthoryear{Kwea}{2022}]{ngppl}
\begin{botherref}
\oauthor{\bsnm{Kwea}}:
ngp\_pl.
GitHub
(2022)
\end{botherref}
\endbibitem

%%% 29
\bibitem[\protect\citeauthoryear{Tang}{2022}]{torch-ngp}
\begin{botherref}
\oauthor{\bsnm{Tang}, \binits{J.}}:
Torch-ngp: a PyTorch implementation of instant-ngp.
https://github.com/ashawkey/torch-ngp
(2022)
\end{botherref}
\endbibitem

%%% 30
\bibitem[\protect\citeauthoryear{Kingma and Ba}{2014}]{adamoptim}
\begin{botherref}
\oauthor{\bsnm{Kingma}, \binits{D.P.}},
\oauthor{\bsnm{Ba}, \binits{J.}}:
Adam: A method for stochastic optimization.
arXiv preprint arXiv:1412.6980
(2014)
\end{botherref}
\endbibitem

%%% 31
\bibitem[\protect\citeauthoryear{Zhang et~al.}{2018}]{LPIPS}
\begin{bchapter}
\bauthor{\bsnm{Zhang}, \binits{R.}},
\bauthor{\bsnm{Isola}, \binits{P.}},
\bauthor{\bsnm{Efros}, \binits{A.A.}},
\bauthor{\bsnm{Shechtman}, \binits{E.}},
\bauthor{\bsnm{Wang}, \binits{O.}}:
\bctitle{The unreasonable effectiveness of deep features as a perceptual metric}.
In: \bbtitle{Proceedings of the IEEE Conference on Computer Vision and Pattern Recognition},
pp. \bfpage{586}--\blpage{595}
(\byear{2018})
\end{bchapter}
\endbibitem

%%% 32
\bibitem[\protect\citeauthoryear{Loshchilov and Hutter}{2017}]{loshchilov2017decoupled}
\begin{botherref}
\oauthor{\bsnm{Loshchilov}, \binits{I.}},
\oauthor{\bsnm{Hutter}, \binits{F.}}:
Decoupled weight decay regularization.
arXiv preprint arXiv:1711.05101
(2017)
\end{botherref}
\endbibitem

\end{thebibliography}
%% if required, the content of .bbl file can be included here once bbl is generated
%%\input sn-article.bbl

\end{document}